%% file: iclr2027_conference.tex
\documentclass{article}

\usepackage{iclr2027_conference,times}

\input{math_commands.tex}

\input{macro}

\usepackage[T1]{fontenc}
\usepackage[utf8]{inputenc}
\usepackage{microtype}
\usepackage{inconsolata}

\usepackage{amsmath}
\usepackage{amssymb}
\usepackage{amsfonts}
\usepackage{nicefrac}

\usepackage{graphicx}
\usepackage{subcaption}
\usepackage{wrapfig}
\usepackage{booktabs}
\usepackage{tabularx}
\usepackage{makecell}
\usepackage{array}
\usepackage{multirow}

\usepackage{algorithm}
\usepackage{algpseudocode}

\usepackage[most]{tcolorbox}
\tcbuselibrary{listings,breakable}
\usepackage{listings}
\usepackage{upquote}
\input{box_format}

\usepackage{xcolor}
\usepackage{hyperref}
\usepackage{url}
\usepackage{cleveref}

\hypersetup{
    colorlinks=true,
    citecolor=customcitecolor,
    linkcolor=magenta,
    urlcolor=blue
}

\Crefname{section}{Section}{Sections}
\Crefname{table}{Table}{Tables}
\crefname{section}{Sec.}{Secs.}
\crefname{table}{Tab.}{Tabs.}
\crefname{figure}{Fig.}{Figs.}

\newcommand{\methodname}{PaperCompiler}
\iclrfinalcopy
\title{\methodname: Faithful Paper-to-Code Generation via Repository-Level Specification Compilation}

\author{Yunhao Liu\\NTU Singapore \And Hong Phuc Pham\\NTU Singapore \And Jaehong Yoon\textsuperscript{\dag}\\NTU Singapore} 
\begin{document}

\maketitle
\renewcommand{\thefootnote}{\fnsymbol{footnote}}
\footnotetext[2]{Corresponding author}
\renewcommand{\thefootnote}{\arabic{footnote}}
\lhead{Preprint.}
\begin{abstract}
Faithfully translating research papers into repository-level implementations remains challenging because papers often describe methods at a high level, leave implementation assumptions implicit, and require generated repositories to preserve method logic, evaluation protocols, and cross-file consistency. 
Despite recent advances in paper-to-code agents, their intermediate outputs are often presented as free-form plans or summaries that downstream coding agents may ignore, reinterpret, or compress, leading to algorithmic simplification and inconsistent repository structure. To address these challenges, we introduce \methodname{}, 
a paper-to-code generation framework that compiles paper-grounded evidence into explicit repository-level implementation specifications.
\methodname{} grounds implementation-relevant evidence while preserving source provenance and distinguishing paper-supported, inferred, externally delegated, and unresolved information. The resulting specifications encode non-degradation requirements, ownership assignments, cross-file dependencies, and file-level constraints. Repository generation proceeds under these compiled specifications while retaining flexibility over local engineering choices not fixed by the paper.
\methodname{} outperforms strong baselines on Paper2CodeBench, achieving a 13.8\% relative improvement in reference-based fidelity (3.64$\rightarrow$4.15) and reducing high-severity evaluator critiques (13.2\%$\rightarrow$6.1\%).

\end{abstract}

\section{Introduction}
Translating machine-learning papers into repository-level implementations is important for reproducibility and practical adoption, yet remains challenging when official code is unavailable or incomplete~\citep{papercoder, paperbench}. Papers often leave implementation-critical details implicit, from data preprocessing and model initialization to evaluation conventions, requiring systems to recover these decisions while maintaining fidelity and consistency throughout the implementation.

\begin{wrapfigure}{rt}{0.46\textwidth}
    \centering
    \vspace{-0.15in}
    \includegraphics[width=\linewidth]{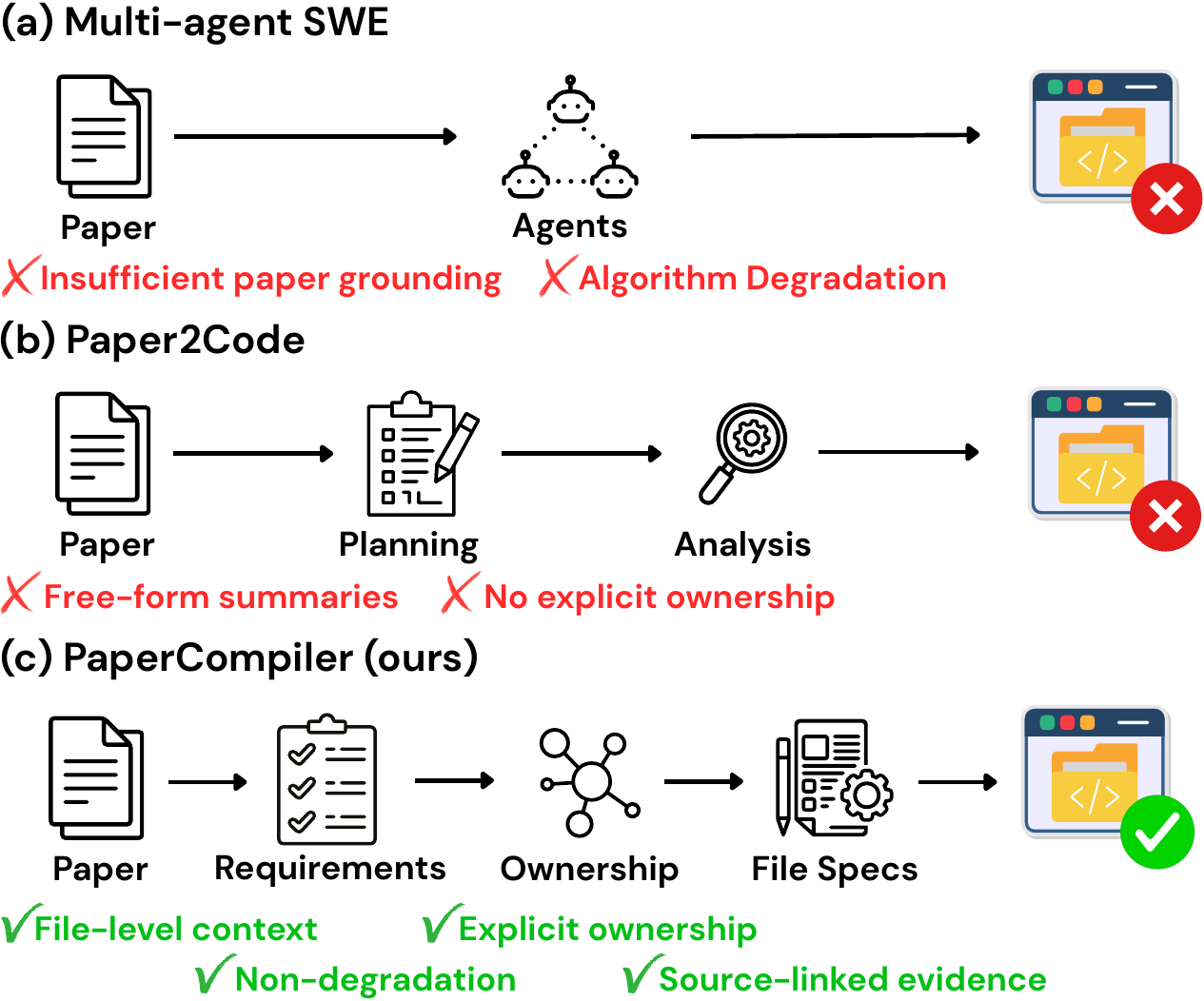}
    \vspace{-0.2in}
    \caption{Comparison of paper-to-code generation workflows.}
    \label{fig:trace-challenges}
    \vspace{-0.35in}
\end{wrapfigure}
Recent code-capable large language models (LLMs) have enabled increasingly automated approaches to this problem~\citep{deepcode, swebench}. General-purpose software-engineering agents can coordinate repository-level development, but are not explicitly designed to recover and preserve paper-specific methodological constraints (\cref{fig:trace-challenges}(a)). Paper-specific systems address this gap through dedicated paper analysis and reproduction workflows: PaperCoder~\citep{papercoder} decomposes repository synthesis into planning, analysis, and coding; AutoP2C~\citep{autoPaperCoder} incorporates multimodal evidence; and AutoReproduce~\citep{autoreproduce} retrieves implicit knowledge from paper lineage and uses generated tests for refinement. However, these systems typically communicate paper-derived knowledge across stages through free-form plans, summaries, or reasoning traces (\cref{fig:trace-challenges}(b)). Such intermediates do not explicitly bind implementation requirements to the repository components responsible for realizing them, allowing critical details to be lost, weakened, or reinterpreted during generation and leading to method-level deviations and cross-file inconsistencies.

To address this remaining bottleneck, we introduce \methodname{}, a paper-to-code generation framework based on specification compilation (\cref{fig:trace-challenges}(c)). 
\methodname{} transforms implementation evidence from the paper into persistent specifications that link each requirement to the repository components responsible for implementing it. This compilation process determines the implementation requirements supported by the available evidence, while keeping uncertain details explicit and mapping each requirement to its implementation location. These specifications maintain method-critical semantics and coordinate cross-file dependencies throughout generation.

\methodname{} consists of three phases: (1) \textit{Paper Grounding} constructs an implementation blueprint while retaining supporting source evidence needed by later stages. (2) \textit{Specification Compilation} reconciles the grounded evidence into explicit implementation requirements and organizes them into repository-level ownership, cross-file dependencies, and localized file specifications, including constraints against method-degrading simplifications. (3) \textit{Constraint-Guided Repository Generation} generates files under the compiled specifications while maintaining interfaces and cross-file consistency. Paper-grounded requirements are carried through these stages to their concrete implementation locations, reducing their weakening or reinterpretation during repository construction.

We evaluate \methodname{} on 90 papers from Paper2CodeBench~\citep{papercoder} and Paper2Code-Extra (P2C-Ex)~\citep{hiras}, comparing it with baselines, including PaperCoder~\citep{papercoder}, AutoP2C~\citep{autoPaperCoder}, and AutoReproduce~\citep{autoreproduce}.
The proposed \methodname{} achieves consistent improvements under matched settings: 4.7\% in reference-free evaluation (4.562$\rightarrow$4.777), 4.3\% under P2C-Ex (4.535$\rightarrow$4.728), and 13.8\% in reference-based fidelity (3.647$\rightarrow$4.152). The gains are largest when generated repositories are compared against author implementations, indicating improved fidelity to paper-specific implementation details beyond surface-level completeness. This is also reflected in the evaluator-error analysis, where \methodname{} reduces the high-severity failure rate (13.2\% $\rightarrow$ 6.1\%).

\section{Related Work}
\paragraph{Paper-to-Code Workflows}
Unlike self-contained program-synthesis tasks~\citep{mbpp,livecodebench}, paper-to-code requires recovering method details and coordinating an entire repository. PaperCoder~\citep{papercoder} separates planning, analysis, and coding; AutoP2C~\citep{autoPaperCoder} adds multimodal extraction and debugging; AutoReproduce~\citep{autoreproduce} leverages paper lineage and execution feedback; and DeepCode combines blueprints, retrieval, memory, and iterative correction~\citep{deepcode}. While these systems improve how implementation knowledge is extracted and refined, information passed between stages is often represented as high-level plans or summaries, leaving downstream generation to reinterpret which details must be preserved and how they map across repository components. PaperCompiler instead compiles paper-grounded evidence into explicit repository-level specifications that preserve implementation requirements and their ownership throughout generation.

\paragraph{Repository-Level Generation and Context Consistency}
Repository-level code generation requires coordinating information distributed across files and dependencies. Prior work~\citep{repocoder,repofusion} improves this through cross-file retrieval or repository-aware training, while RepoBench~\citep{repobench} and CrossCodeEval~\citep{crosscodeeval} evaluate whether models can effectively use such distributed context. CodePlan introduces dependency-aware planning~\citep{codeplan}, while SWE-agent, OpenHands, and Agentless support locating, modifying, and validating components in existing repositories~\citep{sweagent,openhands,agentless,swebench}. These approaches largely assume that the repository structure and interfaces are already available. PaperCompiler instead constructs this organization from the paper, explicitly assigning file ownership, interfaces, and producer--consumer dependencies before code generation.

\paragraph{Evaluation of Paper-to-Code Fidelity}
Existing benchmarks assess different aspects of repository-level coding and research automation. ML-Bench evaluates agents on existing ML repositories, while MLE-bench and RE-Bench emphasize longer-horizon engineering and research tasks~\citep{mlbench,mlebench,rebench}. PaperBench instead targets end-to-end replication of research results, whereas Paper2CodeBench measures repository-level alignment between generated implementations and source papers~\citep{paperbench,papercoder}. Since our focus is faithful translation of paper-specific methods into code, we evaluate on Paper2CodeBench using complementary reference-free, P2C-Ex, and reference-based protocols. In particular, reference-based evaluation compares generated repositories against author implementations, making it more sensitive to methodological deviations that may remain hidden under paper-only evaluation.

\begin{figure*}[t]
\centering
\vspace{-0.1in}
\includegraphics[width=0.9\textwidth]{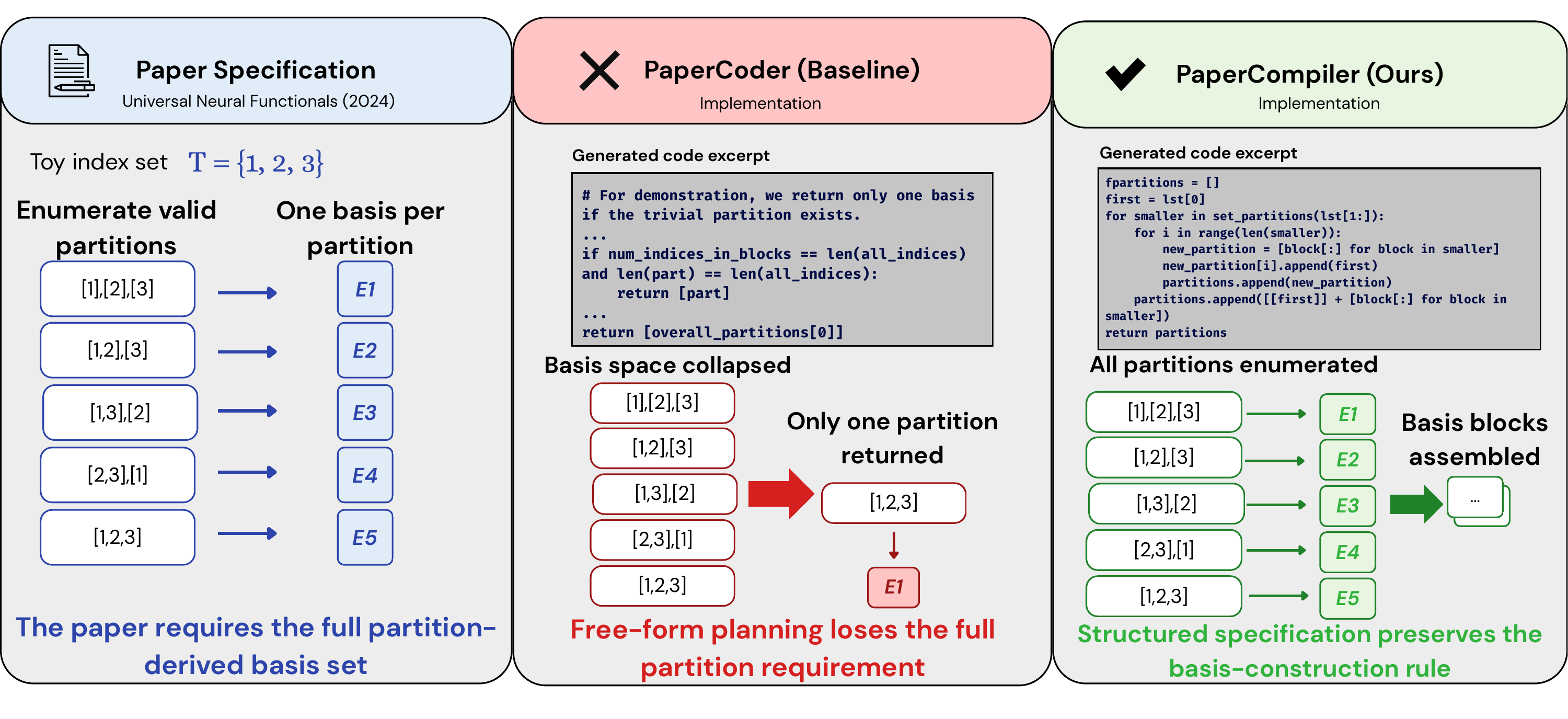}
\vspace{-0.1in}\caption{ 
Case study on \textit{Universal Neural Functionals}.
The left panel uses a simplified toy index set to illustrate Algorithm 1 in \cite{zhou2024universal}: the method constructs one basis element for each valid partition, and the layer is built from the resulting full basis set. The middle and right panels show excerpts from generated code. PaperCoder collapses this requirement by keeping only a single partition candidate, which removes most basis elements from the implementation. In contrast, PaperCompiler preserves the partition-to-basis construction rule and generates code that enumerates valid partitions and assembles the corresponding basis blocks.
}
\label{fig:unf_case_study}
\end{figure*}

\section{Paper-to-Repository Generation via Specification Compilation}
\label{sec:method}

We formulate paper-to-code repository generation as a controlled information transformation problem. Given a machine learning paper $\mathcal{P}$, the goal is to synthesize a repository $\mathcal{C}=\{c_1,\ldots,c_n\}$, where each $c_i$ denotes an implementation file or module, such that the repository collectively implements the paper's main method, data processing, training or execution procedure, and evaluation protocol. The central challenge is not only to generate code, but to preserve implementation-relevant information as the paper is transformed into a coherent multi-file software system.
Paper-grounded facts must remain clearly distinguished from inferred implementation decisions, core methodological requirements must not be diluted into generic approximations, and artifacts produced by one file must be consumed by downstream files with consistent semantics.

As the implementation example shown in \cref{fig:unf_case_study}, existing workflows often compress paper-specific implementation details into coarse intermediate plans, leaving file-level generation to independently resolve missing requirements and thereby causing algorithmic degradation or cross-file inconsistency.To address this information loss, \methodname{} treats repository generation as specification compilation. It separates three questions that existing workflows often conflate: (1) what is supported by the paper, (2) which requirements must be preserved or remain unresolved, and (3) where each requirement belongs in the repository. The resulting specifications link paper-derived requirements to their evidence and implementation owners, while defining cross-file artifact flows and interface semantics before code generation. They therefore constrain both local implementation and cross-file dependencies, reducing the risk of methodological simplification or semantic drift.

Formally, we represent \methodname{} as three conceptual phases:
\begin{align}
(B,Q) &= \mathrm{Ground}(\mathcal{P}), \\
(K,G,\{S_i\}_{i=1}^{n}) &= \mathrm{Compile}(B,Q),
\end{align}
\begin{align}
c_i &= \mathrm{Generate}(f_i,S_i,\mathcal{C}_{<i},\mathcal{S_\text{down(i)}}),
\quad f_i\in \mathrm{TopologicalOrder}(G),\\
\mathcal{C} &= \{c_i\}_{i=1}^{n}.
\end{align}
Here, $\mathrm{Ground}$ denotes \textit{Paper Grounding}, which combines \textit{Blueprint Construction} and \textit{Reference Extraction} to produce an implementation blueprint $B$ and a reference registry $Q$. $\mathrm{Compile}$ denotes \textit{Specification Compilation}, where \textit{Requirement Reconciliation}, \textit{Ownership-Guided Architecture Synthesis}, and \textit{File-Level Contracting} transform the grounded evidence into a reconciled implementation specification $K$, a repository-level ownership graph $G$, and file-level specifications $\{S_i\}_{i=1}^{n}$. Finally, $\mathrm{Generate}$ denotes \textit{Constraint-Guided Repository Generation}, which generates each file $c_i$ from its file-level specification $S_i$, previously generated code $\mathcal{C}_{<i}$ as committed implementation context, and relevant downstream specifications $\mathcal{S}_{\mathrm{down}(i)}$ as compatibility constraints.

\begin{figure*}[t]
    \centering
    \includegraphics[width=0.95\textwidth]{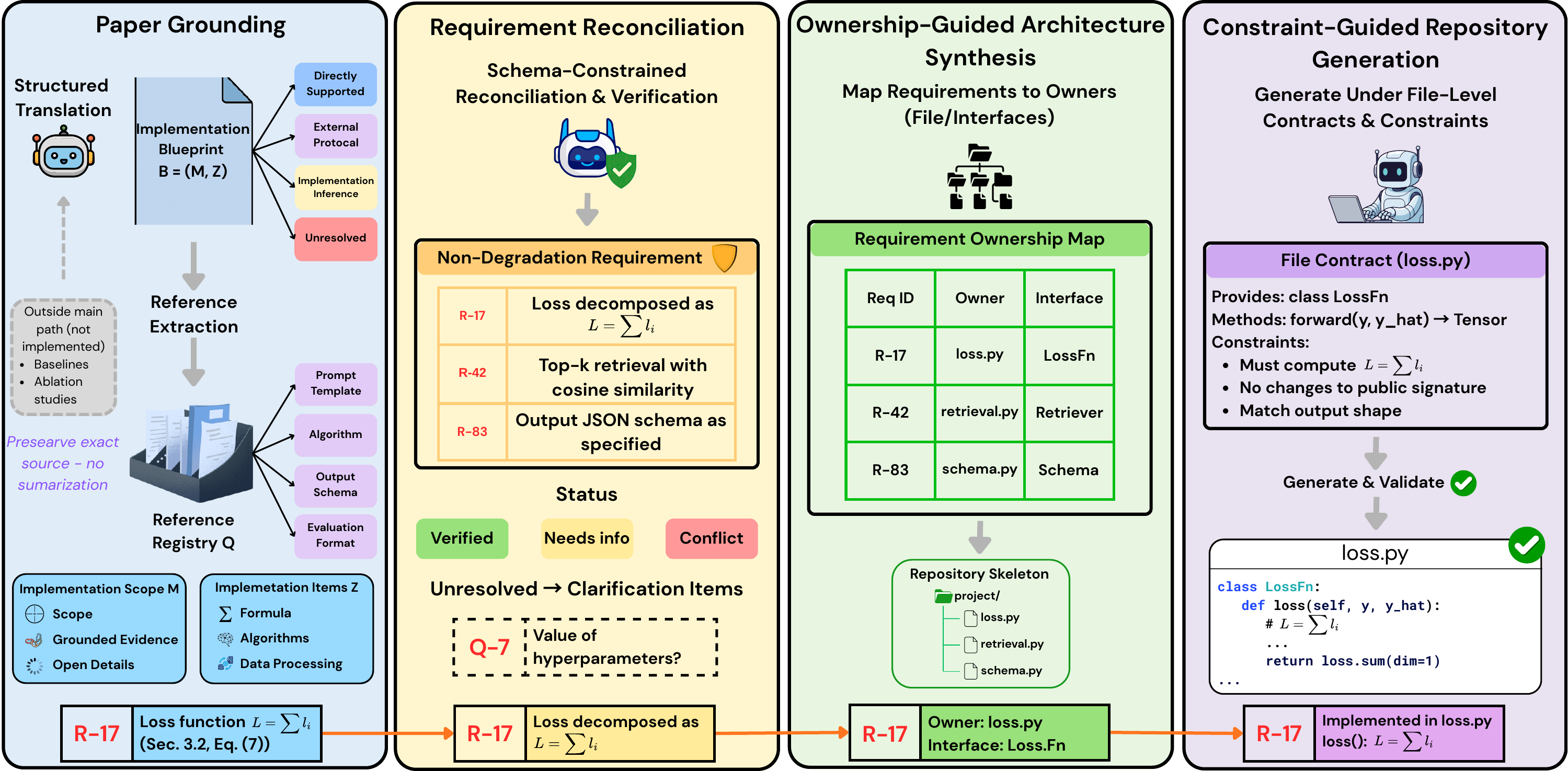}
    \vspace{-0.05in}\caption{Overview of the \methodname{} pipeline. MinerU first converts the paper into Markdown. \textit{Paper Grounding} constructs an implementation blueprint and preserves long or format-sensitive evidence in a reference registry. \textit{Specification Compilation} reconciles grounded evidence into non-degradation requirements, assigns ownership and artifact flows across the repository, and produces localized file-level specifications. \textit{Constraint-Guided Repository Generation} then generates files in dependency order using these specifications, preserved references, committed upstream code, and downstream compatibility constraints. Labels such as \texttt{R-17} and \texttt{Q-7} are illustrative identifiers for a reconciled requirement and a reference-registry item, respectively.}
    \vspace{-0.05in}
    \label{fig:trace-overview}
\end{figure*}

\subsection{Paper Grounding: Blueprint Construction and Reference Extraction}
Paper Grounding converts the parsed paper into structured implementation evidence before the repository architecture is determined, preventing relevant information from being lost prematurely. It consists of \textit{Blueprint Construction}, which constructs a compact implementation blueprint, and \textit{Reference Extraction}, which preserves long or format-sensitive source evidence that should not be compressed into the blueprint.
Blueprint Construction uses a structured LLM prompt with a fixed output schema to identify the main implementation scope, the execution flow from inputs to reported outputs, and key details of the model, data processing, training or execution procedure, and evaluation protocol.
\methodname{} records each atomic implementation item as
$
z=(x,\ell,\tau,r), 
$
where $x$ describes the implementation detail, $\ell$ identifies its source location in the paper, $\tau$ records its evidence status, and $r$ specifies its intended implementation role.
The evidence locator $\ell$ may refer to a section, table, equation, algorithm, appendix, or external reference. 
The evidence-status tag $\tau$ distinguishes paper-supported, externally delegated, inferred, and unresolved items, while $r$ specifies their downstream implementation role, such as a model component, objective, data format, training or evaluation procedure, runtime boundary, or external dependency.

The resulting implementation blueprint is
$
B=(M,Z),
$
where $M$ defines the implementation scope by distinguishing the paper's primary method from baselines, optional analyses, and ablation-only components, and
$Z=\{z_j\}_{j=1}^{m}$ contains the extracted implementation items. This grounded representation provides Specification Compilation with an explicit basis for deciding what must be preserved, what may be inferred, and what should remain unresolved.
Some source material is too long or format-sensitive to retain safely as a compact record, such as prompt templates, output schemas, algorithm listings, or benchmark-specific evaluation formats. For such items, Blueprint Construction issues a reference-extraction request specifying the source location, material type, and downstream use. Reference Extraction copies the requested material from the parsed paper into a reference registry $Q$ without summarizing it. For example, an appendix-defined output schema can be preserved in Q in its original form. These outputs, $B$ and $Q$, provide the grounded evidence used by subsequent Specification Compilation and repository generation.

\subsection{Specification Compilation: From Reconciled Requirements to File-Level Contracts}
\label{sec:spec-compilation}

\textit{Specification Compilation} transforms the grounded blueprint $B=(M,Z)$ and reference registry $Q$ into a reconciled specification $K$, a repository-level ownership graph $G$, and file-level specifications $\{S_i\}_{i=1}^{n}$. It consists of three operations: \textit{Requirement Reconciliation}, \textit{Ownership-Guided Architecture Synthesis}, and \textit{File-Level Contracting}.

\paragraph{Requirement Reconciliation.}
Grounded implementation items preserve what is stated, inferred, externally delegated, or unresolved in the paper, but they do not yet specify how these details should constrain repository generation. Requirement Reconciliation verifies these items against their associated evidence, groups related items into method-level requirements, and preserves their evidence status and provenance. Each reconciled requirement is represented as
\begin{equation}
k=(\mathrm{id},\mathrm{role},\mathrm{src},\mathrm{req},\mathrm{bdry},\mathrm{forbid}),
\end{equation}
where $\mathrm{src}$ links the requirement to grounded evidence, $\mathrm{req}$ specifies the behavior to preserve, $\mathrm{bdry}$ records relevant semantic or runtime boundaries, and $\mathrm{forbid}$ identifies substitutions that would weaken the intended method. The resulting specification
$
K=\{k_j\}_{j=1}^{m_K}
$
covers method behavior, artifact semantics, training or evaluation protocols, external dependencies, and unresolved decisions. It also records abstract producer--consumer relations for artifacts whose semantics must remain consistent across repository components.

\paragraph{Ownership-Guided Architecture Synthesis.}
Architecture Synthesis maps the reconciled requirements in $K$ to a concrete repository design. Let
$F=\{f_1,\ldots,f_n\}$ denote the implementation files and
$K_{\mathrm{core}}\subseteq K$ the requirements on the main method path. We define a primary ownership function
\begin{equation}
\omega:K_{\mathrm{core}}\rightarrow F,
\end{equation}
where $\omega(k)$ identifies the file responsible for implementing or defining requirement $k$. A requirement may still be consumed by other files through public interfaces or shared artifacts.

Let A denote the set of tracked cross-file artifacts, for each tracked artifact $a\in\mathcal{A}$, Architecture Synthesis additionally assigns a producer $\pi(a)\in F$ and consumers $\Gamma(a)\subseteq F$. Together with dependency edges induced by interfaces, imports, and artifact handoffs, these assignments define the repository graph
\begin{equation}
G=(F,\mathcal{E},\omega,\pi,\Gamma).
\end{equation}
Each file is therefore associated with its semantic role, owned requirements, public interfaces, produced or consumed artifacts, and relevant unresolved constraints. The generation dependencies in $\mathcal{E}$ are kept acyclic to support dependency-compatible code generation, without restricting cyclic interactions that may occur at runtime.

\paragraph{File-Level Contracting.}
Contracting localizes the repository-level specification into the information required to generate each file. For file $f_i$, \methodname{} first constructs
\begin{equation}
\mathrm{ctx}_i=\mathrm{Slice}(K,G,Q,f_i),
\end{equation}
where $\mathrm{Slice}$ deterministically selects the requirements, interfaces, artifact relations, dependencies, unresolved cases, and reference materials relevant to $f_i$. This context is compiled into
\begin{equation}
S_i=(I_i,A_i,R_i,H_i,D_i),
\end{equation}
where $I_i$ specifies public interfaces, $A_i$ the implementation recipe, $R_i$ produced and consumed artifacts, $H_i$ cross-file handoff requirements, and $D_i$ non-degradation or unresolved constraints. Thus, $S_i$ localizes the obligations assigned to $f_i$ while retaining the cross-file information necessary for compatibility. Missing or contradictory requirements remain explicit.

\subsection{Constraint-Guided Repository Generation}
\label{sec:repository-generation}

Given the compiled specifications $\mathcal{S}=\{S_i\}_{i=1}^{n}$ and repository graph $G$, \textit{Constraint-Guided Repository Generation} produces implementation files in a dependency-compatible topological order. For each file $f_i$, Engineering uses $S_i$ as the primary generation instruction, previously generated code $\mathcal{C}_{<i}$ as committed implementation context, and relevant downstream specifications $\mathcal{S}_{\mathrm{down}(i)}$ as compatibility constraints.

Previously generated files are treated as committed with respect to their paths, public APIs, schemas, artifact names, and externally visible behavior, while downstream specifications expose the interfaces and artifacts that later files expect to consume. This prevents individual generation steps from independently redefining shared assumptions. When the compiled specification contains an unresolved dependency, unsupported mode, or incompatible boundary, Engineering preserves the specified interfaces and method-specific requirements while making the limitation explicit.

In this way, repository generation remains constrained by obligations compiled from grounded paper evidence, while the LLM retains flexibility over local engineering choices that are not fixed by the paper or repository specification.

\section{Experiments}


\paragraph{Benchmark, baselines, and backbone.}
We evaluate \methodname{} on Paper2CodeBench~\citep{papercoder} using three 30-paper subsets from ICLR, ICML, and NeurIPS 2024, for a total of 90 papers. We compare \methodname{} with general multi-agent software-development baselines, ChatDEV~\citep{chatdev} and MetaGPT~\citep{metagpt}, as well as recent paper-to-code agents, including AutoP2C~\citep{autoPaperCoder}, AutoReproduce~\citep{autoreproduce}, and PaperCoder~\citep{papercoder}. 
{For the latter comparison, all systems receive the same MinerU-parsed Markdown inputs and use o3-mini for generation and o3-mini-high for evaluation, with PaperCoder serving as the most directly comparable staged paper-to-code baseline in this setting.}

Following Paper2CodeBench~\citep{papercoder}, we report reference-free scores using the target paper alone and reference-based scores that additionally consult the author repository when available. We also report P2C-Ex~\citep{hiras}, a finer-grained reference-free protocol. All scores use a 1--5 scale and aggregate multiple independently sampled judge outputs across papers.

\subsection{Main Results}
\label{sec:exp-extended-baselines}

Table~\ref{tab:PaperCoder-batch-main} presents the full-benchmark comparison on Paper2CodeBench~\citep{papercoder}. General multi-agent software-development baselines, ChatDEV and MetaGPT, obtain substantially lower scores and exhibit relatively large standard deviations across conference subsets. PaperCoder provides a considerably stronger paper-to-code baseline, but still falls short of \methodname{} and generally shows higher variance. \methodname{} achieves the highest scores across all conference subsets and evaluation protocols, while maintaining lower standard deviations in most settings. Under matched evaluation conditions, the overall average improves from 4.562 to 4.777 in reference-free evaluation, from 4.535 to 4.728 under P2C-Ex, and from 3.647 to 4.152 in reference-based evaluation. The largest gain appears under reference-based evaluation, with a 13.8\% relative improvement, indicating stronger fidelity to paper-specific implementation details.

\begin{figure*}
    \centering
    \begin{subfigure}[t]{0.32\linewidth}
        \centering
        \includegraphics[width=\linewidth]{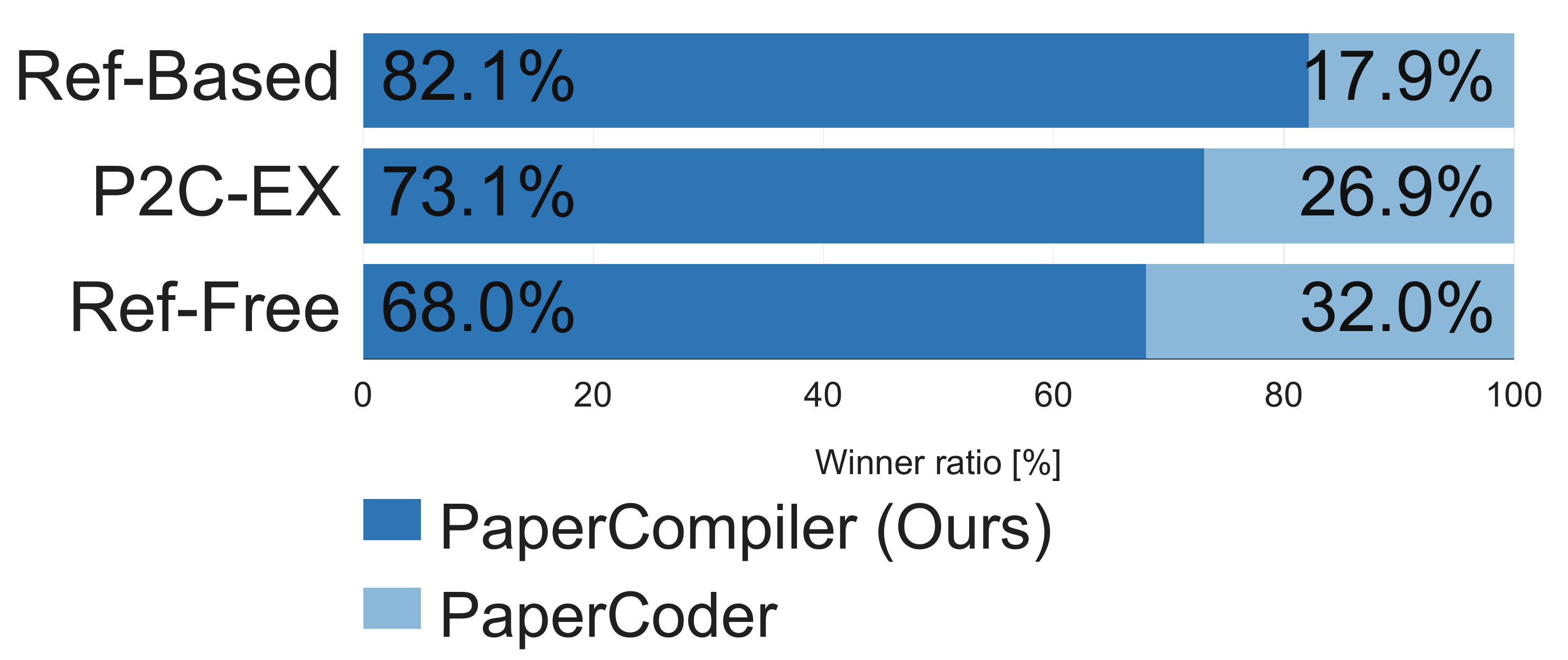}
        \caption{ICLR 2024}
        \label{fig:p2c-winner-ratio-iclr}
    \end{subfigure}
    \hfill
    \begin{subfigure}[t]{0.32\linewidth}
        \centering
        \includegraphics[width=\linewidth]{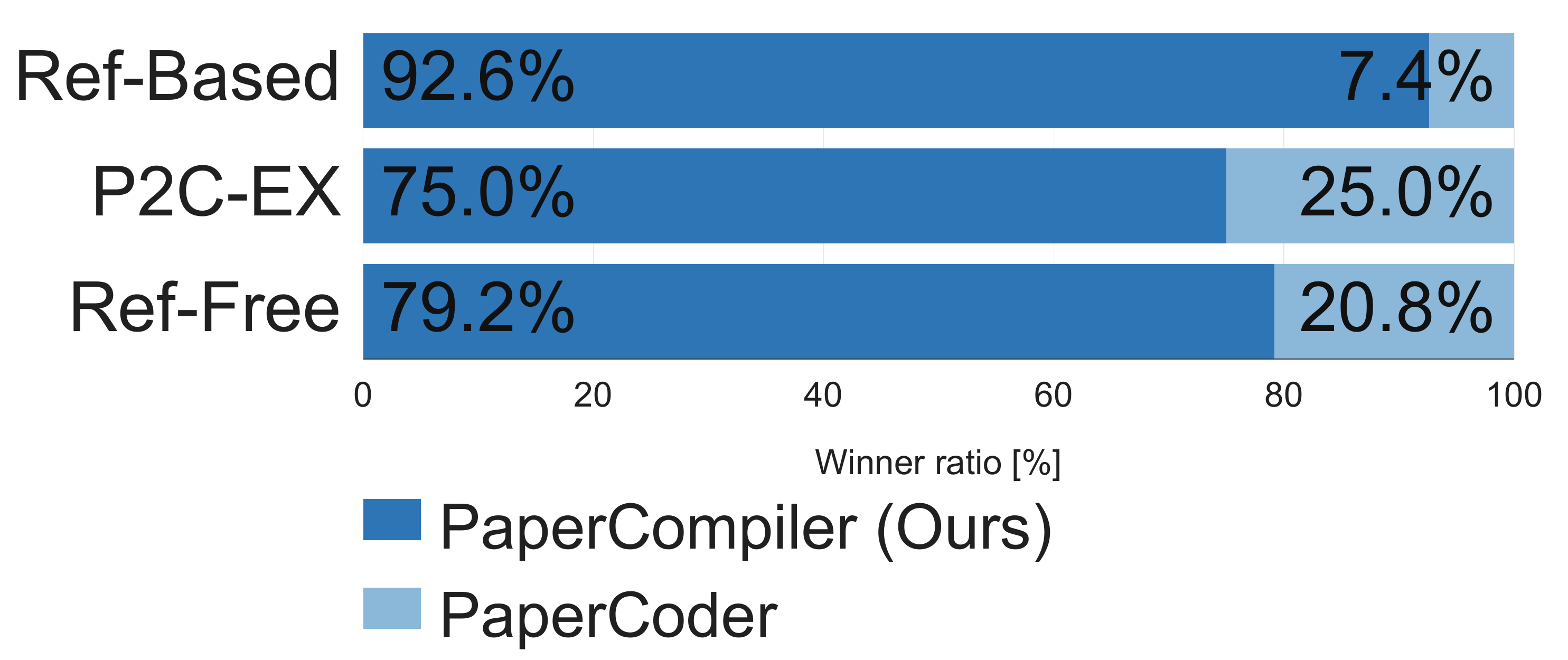}
        \caption{ICML 2024}
        \label{fig:p2c-winner-ratio-icml}
    \end{subfigure}
    \hfill
    \begin{subfigure}[t]{0.32\linewidth}
        \centering
        \includegraphics[width=\linewidth]{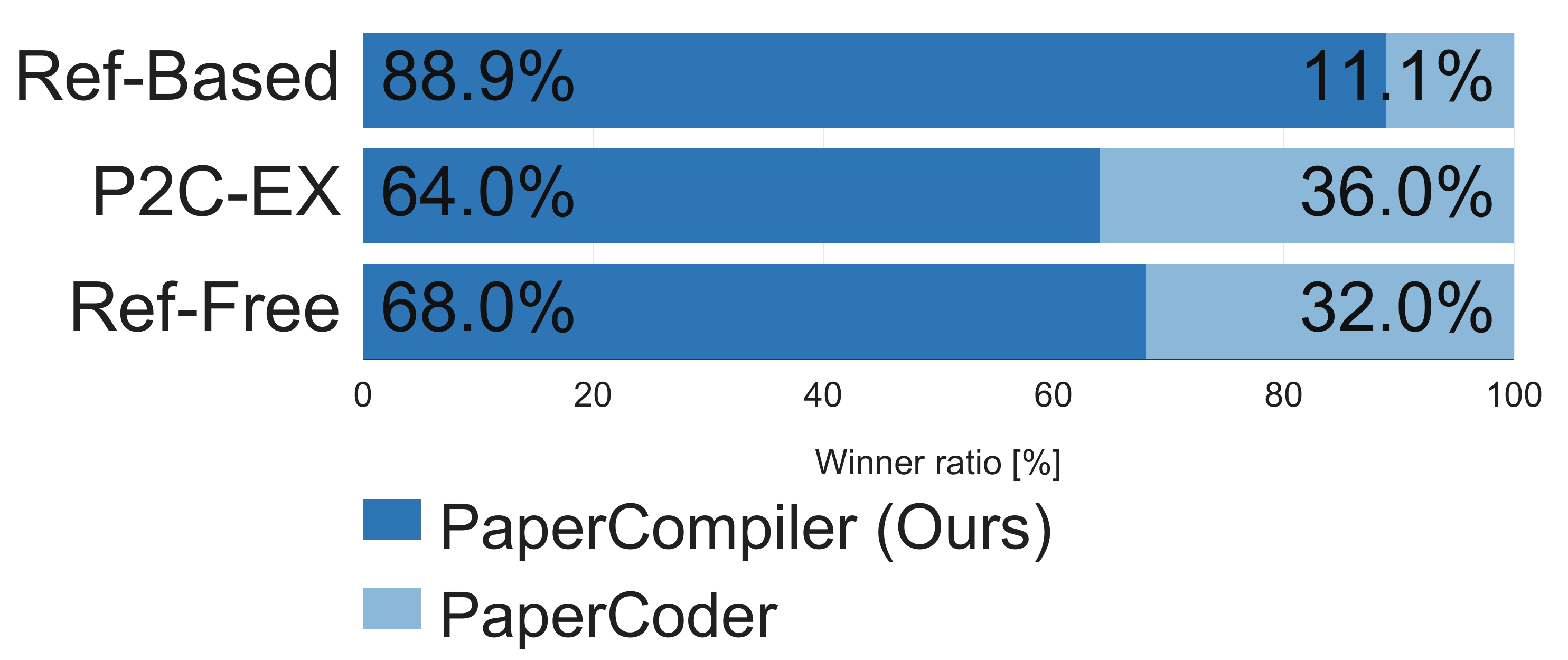}
        \caption{NeurIPS 2024}
        \label{fig:p2c-winner-ratio-neurips}
    \end{subfigure}
    \caption{
    \textbf{Per-paper win ratios between \methodname{} and PaperCoder} on the
    ICLR 2024, ICML 2024, and NeurIPS 2024 subsets of Paper2CodeBench,
    each containing 30 papers. Tie cases are excluded when computing
    the ratios.
    }
    \label{fig:p2c-winner-ratio-by-venue}
\end{figure*}

To assess whether the average gains are broadly distributed across papers, Figure~\ref{fig:p2c-winner-ratio-by-venue} reports per-paper win ratios between \methodname{} and PaperCoder for the ICLR 2024, ICML 2024, and NeurIPS 2024 subsets. Each subset contains 30 papers, with ties excluded from the ratio calculation. \methodname{} attains higher win ratios under all three evaluation protocols across all three subsets. The margin is largest under reference-based evaluation, consistent with the larger average improvement reported in Table~\ref{tab:PaperCoder-batch-main}. The per-paper results therefore show that the gains are broadly distributed rather than being driven by a small number of high-improvement cases.

The gains can be explained by \methodname{}'s explicit specification of paper-grounded requirements. Non-degradation constraints limit method-weakening simplifications, while ownership and cross-file dependencies help maintain consistent implementation semantics across files. This reduces opportunities for paper-specific details to be
lost or reinterpreted during generation.

\begin{table*}[t]
\centering
\scriptsize
\setlength{\tabcolsep}{4.0pt}
\renewcommand{\arraystretch}{1.10}
\resizebox{0.98\linewidth}{!}{\begin{tabular}{ll|lclclc}
\toprule
\multirow{2}{*}{\textbf{Subset}}
& \multirow{2}{*}{\textbf{Method}}
& \multicolumn{2}{c}{\textbf{Reference-free}}
& \multicolumn{2}{c}{\textbf{P2C-Ex}}
& \multicolumn{2}{c}{\textbf{Reference-based}} \\
\cmidrule(lr){3-4} \cmidrule(lr){5-6} \cmidrule(lr){7-8}
& & \textbf{Mean} & \textbf{Std.}
& \textbf{Mean} & \textbf{Std.}
& \textbf{Mean} & \textbf{Std.} \\
\midrule
\multirow{4}{*}{ICLR 2024}
& ChatDEV~\citep{chatdev} & 4.000 & 0.650
& -- & --
& 2.700 & 0.630 \\
& MetaGPT~\citep{metagpt} & 3.520 & 0.600
& -- & --
& 2.480 & 0.480 \\
& PaperCoder~\citep{papercoder}   & 4.500 & 0.504
& 4.479 & 0.536
& 3.667 & 0.566 \\
& \methodname{} (Ours)
& \textbf{4.804} \textcolor{blue}{\scriptsize(+6.8\%)} & \textbf{0.286}
& \textbf{4.771} \textcolor{blue}{\scriptsize(+6.5\%)} & \textbf{0.288}
& \textbf{4.179} \textcolor{blue}{\scriptsize(+14.0\%)} & \textbf{0.406} \\
\midrule
\multirow{4}{*}{NeurIPS 2024}
& ChatDEV~\citep{chatdev} & 4.010 & 0.740
& -- & --
& 2.960 & 0.690 \\
& MetaGPT~\citep{metagpt} & 3.590 & 0.920
& -- & --
& 2.950 & 0.870 \\
& PaperCoder~\citep{papercoder}   & 4.596 & 0.439
& 4.579 & \textbf{0.403}
& 3.617 & 0.601 \\
& \methodname{} (Ours)
& \textbf{4.779} \textcolor{blue}{\scriptsize(+4.0\%)} & \textbf{0.298}
& \textbf{4.700} \textcolor{blue}{\scriptsize(+2.6\%)} & 0.447
& \textbf{4.125} \textcolor{blue}{\scriptsize(+14.0\%)} & \textbf{0.473} \\
\midrule
\multirow{4}{*}{ICML 2024}
& ChatDEV~\citep{chatdev} & 4.120 & 0.530
& -- & --
& 2.970 & 0.580 \\
& MetaGPT~\citep{metagpt} & 3.630 & 0.750
& -- & --
& 2.750 & 0.700 \\
& PaperCoder~\citep{papercoder}   & 4.591 & 0.460
& 4.547 & 0.473
& 3.659 & 0.586 \\
& \methodname{} (Ours)
& \textbf{4.746} \textcolor{blue}{\scriptsize(+3.4\%)} & \textbf{0.456}
& \textbf{4.711} \textcolor{blue}{\scriptsize(+3.6\%)} & \textbf{0.457}
& \textbf{4.151} \textcolor{blue}{\scriptsize(+13.4\%)} & \textbf{0.518} \\
\midrule
\multirow{4}{*}{Average}
& ChatDEV~\citep{chatdev} & 4.043 & --
& -- & --
& 2.877 & -- \\
& MetaGPT~\citep{metagpt} & 3.580 & --
& -- & --
& 2.727 & -- \\
& PaperCoder~\citep{papercoder}   & 4.562 & --
& 4.535 & --
& 3.647 & -- \\
& \methodname{} (Ours)
& \textbf{4.777} \textcolor{blue}{\scriptsize(+4.7\%)} & --
& \textbf{4.728} \textcolor{blue}{\scriptsize(+4.3\%)} & --
& \textbf{4.152} \textcolor{blue}{\scriptsize(+13.8\%)} & -- \\
\bottomrule
\end{tabular}}
\caption{\textbf{Comparison on Paper2CodeBench.} We use the ChatDev and MetaGPT results reported by PaperCoder; PaperCoder and \methodname{} are controlled reruns using the same MinerU Markdown inputs, o3-mini backbone, and o3-mini-high evaluator. Std. denotes standard deviation. P2C-Ex is reported only for the controlled reruns. We highlight relative gains over the strongest baseline in \textcolor{blue}{\textit{blue}}.}
\label{tab:PaperCoder-batch-main}
\end{table*}

\subsection{Comparison with End-to-End Paper-to-Code Systems}
\begin{wrapfigure}{r}{0.50\textwidth}
    \centering
    \vspace{-0.25in}
    \includegraphics[width=\linewidth]{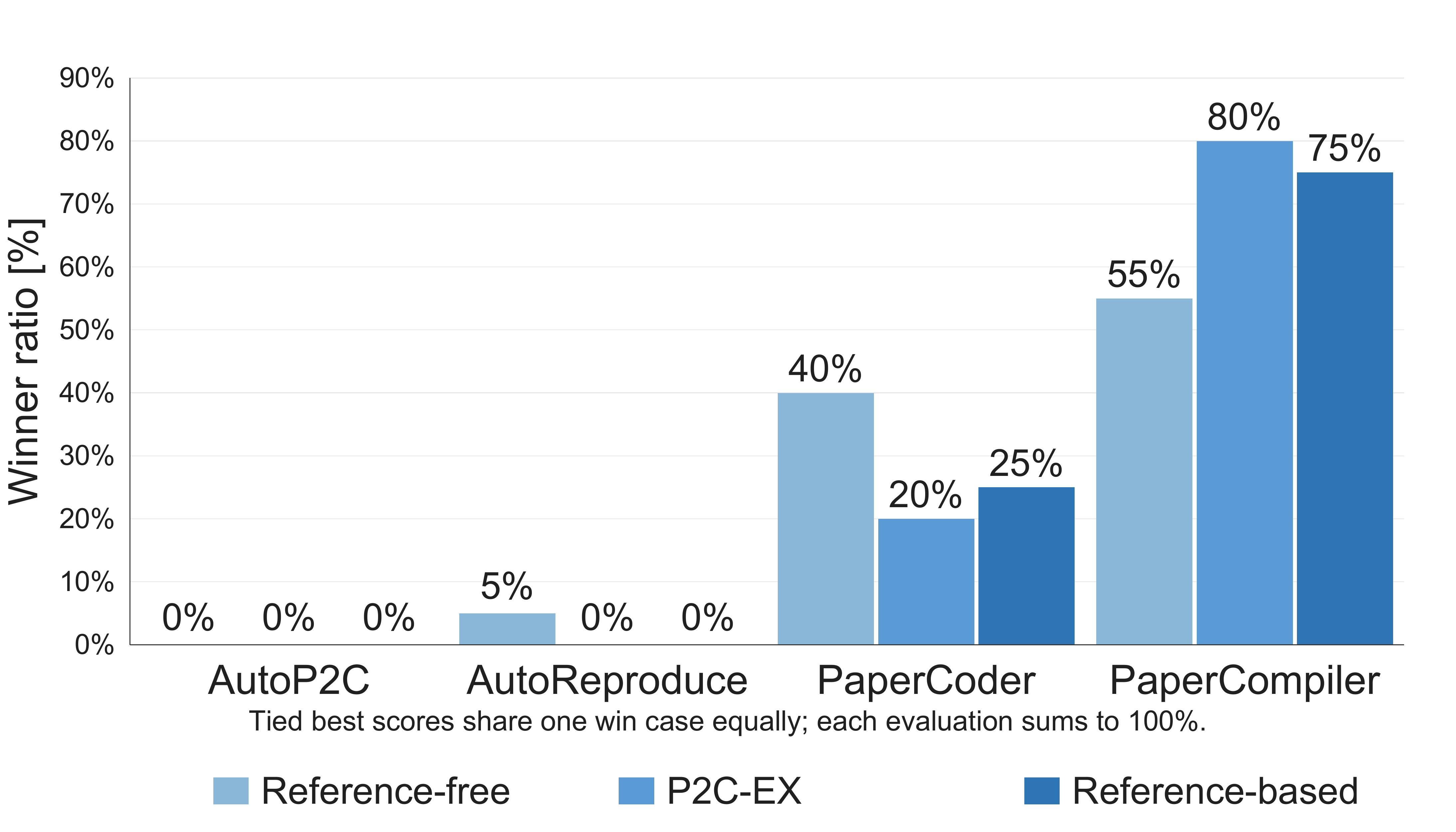}
    \vspace{-0.25in}
    \caption{
    \textbf{Per-paper win ratios} of AutoP2C, AutoReproduce, PaperCoder, and \methodname{} on the ten-paper subset across three evaluation protocols. Ties are divided equally among top-scoring methods.
    }\vspace{-0.25in}
    \label{fig:p2c-10paper-winner-ratio}
\end{wrapfigure}
We further broaden the comparison on the randomly sampled ten-paper subset to include AutoP2C and AutoReproduce alongside PaperCoder (See Appendix for the list of sampled papers). As shown in Table~\ref{tab:extended-baseline-subset}, \methodname{} achieves the highest average score under all three evaluation protocols, outperforming AutoP2C, AutoReproduce, and PaperCoder. The same pattern appears at the paper level in Figure~\ref{fig:p2c-10paper-winner-ratio}: \methodname{} attains the highest per-paper win rate under each protocol, with particularly large margins under P2C-Ex and reference-based evaluation.

AutoP2C incorporates multimodal paper evidence and iterative debugging, while AutoReproduce augments reproduction with paper-lineage knowledge and generated execution tests. These mechanisms provide broader sources of implementation information and validation, but do not consistently preserve paper-specific requirements across repository components. On this subset, both methods remain substantially below \methodname{}, particularly under reference-based evaluation, where alignment with the author implementation is assessed directly. AutoReproduce does not report aggregated token usage because it reports average monetary cost but does not provide aggregate input/output token counts, preventing a directly comparable token estimate.

\methodname{} uses 1.71M tokens per repository on average, compared with 0.98M for PaperCoder. This additional computation reflects the explicit grounding, requirement reconciliation, ownership assignment, and file-level specification steps used to carry paper-specific constraints into repository generation. The performance difference is not explained by token volume alone: AutoP2C uses a comparable 1.68M tokens, yet achieves substantially lower scores across all three protocols. Increasing PaperCoder's generation budget alone would not introduce the explicit requirement reconciliation, ownership structure, or file-level constraints provided by \methodname{}. 

At current o3-mini API rates, the average 1.71M-token usage corresponds to approximately \textbf{\$1.88--\$7.51} per generated repository, depending on the input-output token ratio.
This represents a modest absolute cost for the additional specification-compilation steps relative to the observed fidelity gains.

\begin{table}[t]
\centering
\small
\setlength{\tabcolsep}{5.5pt}
\renewcommand{\arraystretch}{1.10}
\resizebox{0.8\linewidth}{!}{\begin{tabular}{lcccc}
\toprule
\textbf{Method}
& \textbf{Ref-free}
& \textbf{P2C-Ex}
& \textbf{Ref-based}
& \textbf{Avg. tokens} \\
\midrule
AutoP2C~\citep{autoPaperCoder} & 2.700 & 2.700 & 2.088 & 1.68 M \\
AutoReproduce~\citep{autoreproduce} & 3.375 & 3.013 & 2.650 & -- \\
PaperCoder~\citep{papercoder}     & 4.450 & 4.513 & 3.825 & 0.98 M \\
\methodname{} (Ours)   & \textbf{4.813} & \textbf{4.850} & \textbf{4.263} & 1.71 M \\
\bottomrule
\end{tabular}}
\caption{\textbf{Performance-Efficiency comparison} on the Paper2CodeBench subset. Avg. tokens denotes average total tokens per generated repository; ``--'' indicates that aggregated results are unavailable.}
\label{tab:extended-baseline-subset}
\end{table}

\subsection{Semantic and Context-Consistency Failure Analysis}
\label{sec:exp-failure-analysis}

We analyze reference-based critiques from the 90-paper controlled reruns, as comparisons with author repositories are most sensitive to implementation fidelity. Each critique item is manually assigned one primary label from a fixed taxonomy covering algorithmic degradation, missing core components, evaluation mismatch, API/schema mismatch, external-protocol hallucination, misplaced responsibility, producer-consumer break, config-only fake support, and no clear failure. We retain the evaluator-provided severity and report label frequencies for each method.

Table~\ref{tab:failure-analysis} reports the five most frequent failure categories and severity distributions across around 2.6K critique items for each PaperCoder and \methodname{}. Our proposed approach reduces algorithmic degradation from 28.0\% to 24.6\%, with larger reductions in missing core components (12.3\%$\rightarrow$6.8\%) and evaluation mismatches (13.4\%$\rightarrow$8.4\%). High-severity critiques decrease from 13.2\% to 6.1\%, while high-or-medium-severity critiques decrease from 54.2\% to 37.9\%. The no-clear-failure category also increases from 6.7\% to 13.5\%, although this label is sensitive to evaluator phrasing. API/schema mismatches increase modestly from 2.3\% to 4.0\%.

The increase in API/schema mismatches highlights a remaining interface-alignment challenge. Improving fine-grained API consistency is an important direction for future work.

\begin{table}[t]
\centering
\small
\setlength{\tabcolsep}{4pt}
\renewcommand{\arraystretch}{1.08}
\resizebox{0.85\linewidth}{!}{\begin{tabular}{lcc}
\toprule
\textbf{Critique signal} & \textbf{PaperCoder ($n=2561$)} & \textbf{\methodname{} ($n=2649$)} \\
\midrule
Algorithmic degradation & 716 / 28.0\% & 651 / 24.6\% \\
Missing core component & 315 / 12.3\% & 180 / 6.8\% \\
Evaluation mismatch & 343 / 13.4\% & 222 / 8.4\% \\
API/schema mismatch & 59 / 2.3\% & 106 / 4.0\% \\
External protocol hallucination & 76 / 3.0\% & 99 / 3.7\% \\
\midrule
High severity & 337 / 13.2\% & 162 / 6.1\% \\
High or medium severity & 1389 / 54.2\% & 1004 / 37.9\% \\
\bottomrule
\end{tabular}}
\caption{
\textbf{Diagnostic analysis of reference-based evaluator critiques.} Percentages are computed over PaperCoder and \methodname{} critique items, each assigned one primary failure label.
}
\vspace{-0.1in}
\label{tab:failure-analysis}
\end{table}

\subsection{Ablation Study and Qualitative Analysis}
\label{sec:ablation}

\paragraph{Setup.}

We ablate three key components of the proposed approach while regenerating each repository end-to-end. \emph{w/o Reconciliation} removes the non-degradation and method-level requirements before architecture design (\S\ref{sec:spec-compilation}); \emph{w/o Context Slicing} gives every file the full reconciled specification and ownership graph; and \emph{w/o Contracting} generates files without file-level specifications. Blueprint Construction, Ownership-Guided Architecture Synthesis, and Constraint-Guided Repository Generation remain fixed, and all variants reuse upstream artifacts where possible to isolate the targeted mechanism. We randomly sampled nine papers, including \textit{INTL, AutoVP, TransformerCompression, RECOMBINER, WassersteinSSL, INTR, Auto-J, VONet,} and \textit{VDC}. Each paper--variant pair is generated once with \texttt{o3-mini}, yielding 27 ablated repositories plus nine \methodname{} repositories. We use the same three 1--5 evaluation protocols and average eight independent judgments per repository. 


\begin{wraptable}{r}{0.48\textwidth}
\centering
\small
\vspace{-0.15in}
\setlength{\tabcolsep}{6pt}
\renewcommand{\arraystretch}{1.05}
\resizebox{\linewidth}{!}{\begin{tabular}{lccc}
\toprule
Variant & Ref-free & P2C-Ex & Ref-based \\
\midrule
\methodname{}      & 4.57 & 4.42 & \textbf{4.38} \\
w/o Reconciliation     & 4.44 & 4.39 & 3.86 \\
w/o Context Slicing    & \textbf{4.72} & \textbf{4.74} & 4.11 \\
w/o Contracting        & 4.65 & 4.54 & 3.92 \\
\bottomrule
\end{tabular}}
\vspace{-0.05in}
\caption{\textbf{Ablation under three model-based metrics} (correctness, $1$--$5$, with each cell averaging $8$ judge samples per paper).}
\vspace{-0.1in}
\label{tab:ablation}
\end{wraptable}
\textbf{Reference-based evaluation provides a more discriminative signal.}
As shown in Table~\ref{tab:ablation}, the ablations often match or exceed the complete \methodname{} under \emph{ref-free} and \emph{ref-free-ex}, where scores cluster near the top of the scale. Removing Context Slicing, for example, improves these scores by $+0.15$ and $+0.32$ despite reducing \emph{ref-based} fidelity by $0.26$. Reference-free evaluation can reward surface-complete repositories without verifying alignment with the author's implementation. We therefore focus on reference-based results, where \methodname{} outperforms all ablations.

\textbf{Reconciliation has the largest average effect.}
Removing Reconciliation reduces reference-based performance from 4.38 to 3.86 ($-0.51$), followed closely by removing Contracting 3.92 ($-0.46$); removing Context Slicing has a smaller effect 4.11 ($-0.26$). The larger drops from Reconciliation and Contracting highlight the importance of establishing method-level obligations and carrying them into file-level specifications.

\textbf{Context Slicing trades broader context for targeted file-level information.}
Although it has the smallest average reference-based effect (4.38$\rightarrow$4.11), removing Context Slicing on \textit{VDC} replaces multimodal inference with a constant response and reduces performance from $4.75$ to $3.75$.
On \textit{WassersteinSSL} and \textit{INTR}, the unsliced variant performs better, indicating that its benefit can be smaller when broader context already provides useful implementation details. Its main benefit therefore lies in protecting file generation from missing critical dependencies, rather than uniformly improving every repository.

\textbf{Repository-level inspection shows how the ablated components fail in practice.}
\textit{VDC} without Contracting, the multimodal inference module becomes an \texttt{InstructBlipModel} placeholder with an unimplemented forward path, and the pipeline falls back to generic questions. Full \methodname{} instead preserves a functional Instruct-BLIP interface, yielding a 2.25-point reference-based advantage. In \textit{INTR} without Reconciliation, the decoder retains only cross-attention while omitting the self-attention and feed-forward refinement required by the method. These cases show that Contracting helps carry concrete implementation obligations into individual files, while Reconciliation prevents method-level requirements from being weakened during generation.

\section{Conclusion}
\label{sec:conclusion}

We introduce PaperCompiler, a repository-level constraint-guided framework for faithful paper-to-code synthesis. PaperCompiler treats repository generation as a controlled transformation from scientific evidence to implementation specifications, preserving paper-derived requirements, assigning file-level responsibilities, and maintaining cross-file consistency. Experiments on Paper2CodeBench show consistent gains over a strong staged baseline across reference-free, P2C-Ex, and reference-based evaluation, with the largest gain in reference-based fidelity. Ablations identify Reconciliation and Contracting as the most important components, while diagnostic analysis shows fewer severe semantic and method-completeness failures. Overall, reliable paper-to-code generation requires not only stronger LLMs, but also explicit control over how paper-derived information is represented, propagated, and consumed during repository construction.

\subsection*{AI use statement}
Generative AI tools were used only for language polishing and improving the clarity and readability of the manuscript. The authors developed and verified all research methodology, experiments, analyses, results, and conclusions and take full responsibility for the final content.

\bibliography{iclr2027_conference}
\bibliographystyle{iclr2027_conference}

\appendix
\section{Limitations}
\methodname{} currently relies primarily on text-parsed paper content, limiting its ability to recover implementation-critical information conveyed through architecture diagrams, complex figures, visual examples, or other non-textual forms. When such details are absent from the parsed content, the system may still rely on the underlying LLM's parametric knowledge to resolve implicit implementation gaps. This can be challenging for papers that depend on specialized external tools, proprietary APIs, simulators, undocumented benchmark conventions, or auxiliary resources unavailable to the model. Future work could incorporate multimodal document understanding, targeted retrieval, and executable tool interaction to address these cases.

Our evaluation focuses on repository-level method fidelity under controlled paper-to-code protocols and does not establish full reproduction of reported experimental results or executability across all generated repositories, generation seeds, and model families. Moreover, \methodname{}'s requirements and file-level specifications serve as generation-time guidance rather than formal correctness guarantees. Integrating automated validation, execution-based testing, and independent or human verification would provide stronger guarantees in future work.

\section{Extended Experimental Details} For  Table~\ref{tab:extended-baseline-subset} and Figure~\ref{fig:p2c-10paper-winner-ratio}, we evaluate PaperCoder, AutoP2C, AutoReproduce, and \methodname{} on a randomly sampled ten-paper subset of Paper2CodeBench: INTL, AutoVP, TransformerCompression, RECOMBINER, CAML, CARE, GGS, SEABO, iTransformer, and SparseFormer. 

\section{PaperCompiler Algorithm Overview}
Algorithm~\ref{alg:trace} summarizes the end-to-end repository synthesis procedure of \methodname{}. The algorithm follows the three conceptual phases introduced in Section~\ref{sec:method}: \textit{Paper Grounding}, \textit{Specification Compilation}, and \textit{Constraint-Guided Repository Generation}. In Paper Grounding, $T$ denotes \textit{Blueprint Construction}, which extracts the implementation blueprint $B$, while $\mathrm{ExtractReferences}$ preserves requested source materials in the reference registry $Q$. During Specification Compilation, $R$ performs \textit{Requirement Reconciliation} to construct $K$, and $A$ performs \textit{Ownership-Guided Architecture Synthesis} to construct the repository graph $G$. $\mathrm{Slice}$ then extracts the context relevant to each file, and $\mathrm{FileSpec}$ performs \textit{File-Level Contracting} to produce $S_i$. Finally, files are generated in the topological order induced by $G$, so that each generation step can follow its local specification while respecting already committed interfaces and cross-file compatibility constraints.

\begin{algorithm}[H]
\small
\caption{\methodname{} repository synthesis}
\label{alg:trace}
\begin{algorithmic}[1]
\Require paper $\mathcal{P}$
\Ensure generated repository $\mathcal{C}$

\Statex \textbf{Paper Grounding}
\State $B \leftarrow T(\mathcal{P})$ \Comment{construct implementation blueprint}
\State $Q \leftarrow \mathrm{ExtractReferences}(B,\mathcal{P})$ \Comment{preserve requested source evidence}

\Statex \textbf{Specification Compilation}
\State $K \leftarrow R(B,Q)$ \Comment{build reconciled implementation specification}
\State $G \leftarrow A(K)$ \Comment{build repository ownership graph}
\State $\mathcal{S}\leftarrow \emptyset$
\For{$f_i \in F(G)$}
\State $\mathrm{ctx}_i \leftarrow \mathrm{Slice}(K,G,Q,f_i)$
\State $S_i \leftarrow \mathrm{FileSpec}(f_i,\mathrm{ctx}_i)$
\State $\mathcal{S}\leftarrow \mathcal{S}\cup \{S_i\}$
\EndFor

\Statex \textbf{Constraint-Guided Repository Generation}
\State $\mathcal{C}\leftarrow \emptyset$
\For{$f_i \in \mathrm{TopologicalOrder}(G)$}
\State $c_i \leftarrow \mathrm{Engineer}(f_i,S_i,\mathcal{C}_{<i},\mathcal S_{\mathrm{down}(i)})$
\State $\mathcal{C}\leftarrow \mathcal{C}\cup \{c_i\}$
\EndFor
\State \Return $\mathcal{C}$
\end{algorithmic}
\end{algorithm}

\section{Additional Qualitative Case Details}
\label{app:case_details}

This appendix provides additional qualitative diagnostics for selected generated repositories. These cases are not used as statistical evidence; rather, they illustrate how different systems instantiate, weaken, or miss paper-specific implementation structures. We include two main-text cases as evidence notes and two additional appendix cases. The main text uses \textit{universal\_neural\_functional} as a motivating example of algorithmic-detail collapse and \textit{iTransformer} as a compact repository-structure diagnostic. Here we provide the supporting evidence trail and add two further cases: \textit{INTL}, where \methodname{} better preserves a training pipeline, and \textit{SEABO}, where \methodname{} exhibits a remaining failure in protocol and runner integration.

\begin{table}[t]
\centering
\scriptsize
\setlength{\tabcolsep}{4pt}
\renewcommand{\arraystretch}{1.12}
\newcommand{\scorecell}[4]{\makecell[c]{#1\\#2\\#3\\#4}}
\begin{tabular}{@{}l>{\raggedright\arraybackslash}p{2.35cm}cccc@{}}
\toprule
Case & Role & AutoP2C & AutoReproduce & P2C & \methodname{} \\
\midrule
\textit{Universal Neural Functionals}
& Algorithmic basis construction
& \scorecell{--}{--}{--}{--}
& \scorecell{--}{--}{--}{--}
& \scorecell{3.500}{3.500}{2.875}{3.292}
& \scorecell{4.750}{5.000}{4.875}{4.875} \\
\midrule
\textit{iTransformer}
& Repository structure
& \scorecell{4.625}{4.625}{3.125}{4.125}
& \scorecell{5.000}{4.125}{3.125}{4.083}
& \scorecell{5.000}{5.000}{3.875}{4.625}
& \scorecell{4.875}{5.000}{3.625}{4.500} \\
\midrule
\textit{INTL}
& Training pipeline
& \scorecell{2.375}{2.250}{1.750}{2.125}
& \scorecell{4.000}{3.750}{3.625}{3.792}
& \scorecell{3.375}{3.750}{3.000}{3.375}
& \scorecell{5.000}{4.875}{4.500}{4.792} \\
\midrule
\textit{SEABO}
& Failure-oriented diagnostic
& \scorecell{3.875}{4.125}{2.875}{3.625}
& \scorecell{4.000}{4.000}{3.125}{3.708}
& \scorecell{5.000}{5.000}{5.000}{5.000}
& \scorecell{4.625}{4.625}{4.375}{4.542} \\
\bottomrule
\end{tabular}
\caption{
Qualitative case scores. Each score cell reports, from top to bottom, reference-free, P2C-Ex, reference-based, and average scores. Missing entries indicate that no usable generated repository or evaluator output is available for the corresponding system in the artifacts we inspected.
}
\label{tab:qual_case_scores}
\end{table}

\subsection{\textit{Universal Neural Functionals}: algorithmic basis-construction collapse.}
The main-text figure uses \textit{Universal Neural Functionals} as a motivating example of algorithmic degradation. The target implementation structure consists of a specification parser, a UNF layer generator, a layer assembler, a channel-extension component, and training or experiment orchestration. The implementation-critical requirement is that valid partitions should be enumerated and converted into basis elements for the equivariant layer.

The P2C rerun generates a connected implementation path in a flat \texttt{model.py}, including \texttt{EquivariantBasisGenerator} and \texttt{WeightLayer}. However, the partition-generation logic returns only the trivial partition or the first available partition candidate. Thus the generated repository may appear structurally connected while collapsing the basis family to a single candidate. In contrast, \methodname{} separates specification parsing, basis generation, layer assembly, and multi-channel extension into distinct modules. Its generated \texttt{src/unf\_layer\_generator.py} preserves partition enumeration and constructs corresponding basis functions, which are then consumed by the layer assembly logic.

The intermediate artifacts of \methodname{} route this requirement into file-level generation. The specification parser is assigned to a dedicated file, and the basis-generation requirement is routed to \texttt{src/unf\_layer\_generator.py}. The relevant file-level specification forbids hard-coded or heuristic defaults for the specification and preserves the valid-partition mapping needed by the basis generator. Evaluator critiques support the same interpretation: P2C is criticized for returning only a trivial or first partition, whereas \methodname{} is credited for enumerating valid partitions and constructing the corresponding basis representation. This case supports the claim that \methodname{} can preserve a discrete algorithmic requirement through repository generation; it does not claim full end-to-end fidelity for arbitrary high-order weight spaces or all training settings.

\subsection{\textit{iTransformer}: repository-structure diagnostic.}
The main text uses \textit{iTransformer} to illustrate repository-structure behavior rather than a clear win for \methodname{}. The reference implementation is organized as an experiment-oriented forecasting repository with separate components for data providers, models, experiments, utilities, scripts, and experiment entry points. AutoP2C generates a layered application-style repository with configuration, data, model, training, evaluation, and main-entry modules. This preserves the broad variate-as-token idea, but its reference-based evaluation indicates incomplete preservation of the full experimental protocol. 

AutoReproduce compresses the reproduction into a single \texttt{main.py} while preserving the central variate-as-token transformation and a connected training and evaluation path. However, it simplifies the surrounding data pipeline and experimental framework relative to the reference implementation. Its strong reference-free score but lower reference-based score therefore reflects a recurring pattern: the headline method is visible, while repository structure and experimental coverage are only partially recovered.

P2C obtains the strongest score on this example. It preserves the input transposition and Transformer backbone, but compresses the repository into six flat files: configuration, dataset loading, model, trainer, evaluation, and main entry. Its data pipeline is also simplified to a CSV sliding-window setting, omitting part of the multi-dataset experimental structure and advanced attention variants. \methodname{} instead produces a separated \texttt{src/} pipeline with modules for configuration, data loading, variate embedding, Transformer blocks, projection, training, evaluation, and orchestration. Its intermediate artifacts route configuration objects, forecast tensors, transposed tensors, and training or evaluation outputs across these modules. Evaluator feedback indicates that these modules are connected and that inverted tokenization is preserved, but also criticizes \methodname{} for compressing the broader experiment suite into a single \texttt{src/} pipeline and omitting part of the partial-variate training and variate-generalization protocol.

This case illustrates both a benefit and a limitation of specification-guided generation. \methodname{} makes module boundaries and data flow more explicit, but it can still simplify broad experimental protocols when they are not fully grounded or routed. Therefore, the case does not show that the generated directory structure is superior to the original experiment repository; rather, it shows how different systems trade off model-path preservation, repository organization, and experimental coverage.

\subsection{\textit{INTL}: Training-Pipeline Preservation}

\textit{INTL} is included as an appendix case where \methodname{} more clearly improves training-pipeline fidelity. The target implementation requires an image self-supervised training loop that connects augmentation, backbone and projection modules, IterNorm channel whitening, composite losses, training orchestration, and k-NN evaluation. In particular, the IterNorm transform should operate as channel whitening and should be consumed by the training objective.

AutoP2C defines a spectral transformation component with an \texttt{iter\_norm} operation, but evaluator feedback indicates that this component is not properly connected to the projection head or training loop. Thus a method-specific component is present by name but disconnected from the main producer--consumer chain. P2C implements IterNorm in \texttt{model.py}, but its whitening operation centers along the sample dimension, producing a covariance structure over the batch axis. This is identified as an algorithmic degradation.

AutoReproduce produces a connected single-file pipeline spanning augmentation, representation learning, iterative normalization, the training objective, and evaluation. The method-specific transform is therefore consumed by the main training path rather than remaining disconnected. However, several mathematical and evaluation details are simplified relative to the reference implementation, so the repository preserves the broad topology without fully recovering the algorithmic and experimental protocol.

\methodname{} separates the IterNorm transform, loss module, and training-loop executor into dedicated files. Its intermediate artifacts route augmented images from data loading into the backbone, assign IterNorm to \texttt{src/iter\_norm\_transform.py}, assign composite losses to \texttt{src/loss\_module.py}, and assign orchestration to \texttt{src/training\_loop\_executor.py}. The generated implementation uses Newton-iteration-based IterNorm and connects the whitening component with the loss and evaluation pipeline. Remaining gaps include incomplete integration of EMA and multi-crop extensions, which are defined as optional extensions. This case supports the claim that file-level routing can help preserve cross-file whitening--loss--trainer dependencies, while not establishing exact reproduction of every training extension or loss-detail variant.

\subsection{\textit{SEABO}: A Failure-Oriented Diagnostic Case}

\textit{SEABO} is included as a failure-oriented diagnostic case for \methodname{}. The target implementation requires a connected offline-RL pipeline: loading D4RL data, extracting expert trajectories, building a nearest-neighbor structure such as a KD-tree, annotating rewards, training an offline RL method such as TD3\_BC or IQL, and evaluating with the appropriate environment and normalized-return protocol.

AutoP2C produces a lower-scoring generated pipeline, but the currently inspected evidence is insufficient to make a precise structure-level claim about its failure. P2C produces the strongest result on this case. Its generated repository connects expert demonstration extraction, KD-tree construction, reward computation, offline-RL training, and evaluation in the main pipeline. It preserves state-only support and achieves a perfect reference-based score, with remaining omissions mainly concerning optional retrieval variants such as Ball-tree or HNSW.

AutoReproduce connects the main SEABO path in a single file, including offline-data loading, expert selection, nearest-neighbor reward annotation, and offline-RL training. This is a substantive implementation rather than a placeholder pipeline. Nevertheless, it narrows the method to one configuration and simplifies parts of the reward, training, and evaluation protocols, which limits its fidelity and experimental coverage relative to the reference repository.

\methodname{} preserves the main modular topology of the method: data loading, KD-tree construction, reward annotation, offline-RL training, evaluation, and orchestration are separated into dedicated modules. Its intermediate artifacts route global configuration fields such as reward parameters, normalization, and the state-only setting through data loading, reward annotation, training, evaluation, and the main runner. The generated main path also explicitly follows the intended sequence of loading data, extracting expert trajectories, building the KD-tree, annotating rewards, training an offline-RL model, and evaluating.

However, this routed structure is not fully realized in the final executable behavior. The generated runner falls back to a dummy dataset when no real D4RL path is provided, and the offline-RL trainer raises \texttt{NotImplementedError} for the \texttt{state\_only=True} regime. Evaluator feedback confirms that the KD-tree reward annotation and modular offline-RL pipeline are present, but criticizes the use of a dummy environment instead of full D4RL integration and identifies the unsupported state-only path as a high-severity gap. This failure shows that \methodname{}'s specifications can preserve the intended topology while still failing to complete external-protocol and runner integration. It supports the limitation that file-level specifications are generation-time guidance.

\section{Stage Prompt Excerpts}

\begin{tracepromptbox}{Prompt excerpt from Reconciliation: contract construction and schema}
PRIMARY PURPOSE:
Convert the natural-language implementation blueprint into compact, enforceable implementation contracts.

The Alignment Spec must:

1. Verify the planning blueprint against the paper reference.
2. Correct planning drift when the paper evidence disagrees or when planning is too vague for implementation.
3. Preserve the paper's main execution flow from raw input to final reported output.
4. Define core path contracts that routing/coding must not violate.
5. Define object/tensor/state/prompt/artifact flow with producer-consumer semantics.
6. Define runtime boundaries such as layout, device, preprocessing order, prompt-target boundary, checkpoint/state contents, and metric parsing.
7. Define formula and algorithm exactness requirements when implementation details affect correctness.
8. Isolate missing details and implementation choices from paper facts.
9. Keep optional, ablation-only, and baseline-only components out of the main method path.

FACT CLASSIFICATION:
Every implementation-relevant claim must use one of:

* paper_fact:
  Explicitly supported by paper text, table, algorithm, figure, formula, prompt template, or appendix.

* external_contract:
  Explicitly delegated by the paper to another protocol, repository, benchmark, prior paper, API, simulator, dataset convention, or official implementation.

* implementation_choice:
  Not provided by the paper but useful or necessary for runnable code.

* not_applicable:
  Not relevant to this paper's task archetype.

OUTPUT JSON STRUCTURE:

{{
"metadata": {{
"task_type": "...",
"implementation_archetype": "...",
"primary_reproduction_target": "...",
"reproduction_modes": [],
"scope_partition": {{
"main_method": "...",
"required_auxiliary_components": [],
"optional_extensions": [],
"ablations_or_analysis": [],
"baselines_or_external_methods": []
}}
}},

"planning_corrections": [],

"evidence_index": [],

"workflow_contract": {{
"workflow_archetype": {{
"primary": "...",
"secondary": [],
"custom_name_if_needed": ""
}},
"main_execution_path": [],
"lifecycle_edges": [],
"core_artifacts": [],
"external_protocol_required": [],
"forbidden_substitutions": []
}},

"core_path_contracts": [],
"experiment_protocol": {{}},
"execution_flow": [],
"object_contracts": {{}},
"runtime_boundary_contracts": [],
"formula_exactness_contracts": [],
"method_graph": {{
"representation_type": "...",
"objects": {{}},
"nodes": []
}},
"training_or_execution_contract": {{}},
"evaluation_contract": {{}},
"module_alignment_hints": [],
"open_design_choices": [],
"integrity_checks": [],
"compact_handoff": {{
"must_preserve": [],
"must_assign_in_routing": [],
"high_risk_boundaries": [],
"fail_fast_conditions": []
}}
}}
\end{tracepromptbox}

\begin{tracepromptbox}{Prompt excerpt from Architecture: ownership, APIs, and dependency routing}
STAGE-3 PURPOSE:
Assign Stage-2 contracts, objects, execution steps, runtime boundaries, formula contracts, external protocols, experiment fields, open design choices, and integrity checks to concrete implementation files and public APIs.

Your output must enable later stages to:

* generate files in a safe dependency order,
* avoid duplicated or missing logic,
* avoid circular imports,
* keep files at reasonable size,
* preserve every core contract,
* ensure produced artifacts are consumed,
* keep external protocols explicit,
* fail fast rather than silently using unsupported placeholders.

CORE PRINCIPLE:
Design files from Stage-2 contracts outward.
A file exists because it owns a coherent set of Stage-2 contracts, artifacts, APIs, runtime boundaries, formulas, external protocols, or orchestration responsibilities.

ROUTING RESPONSIBILITIES:

1. Define a practical file tree.
2. Assign every Stage-2 core_path_contract to exactly one primary owner.
3. Assign every Stage-2 execution_flow step to exactly one primary owner unless Stage 2 marks it optional.
4. Assign every Stage-2 method_graph node to exactly one primary owner unless optional.
5. Assign every object/artifact with cross-stage use to a producer and one or more consumers.
6. Assign every runtime_boundary_contract to an owner.
7. Assign every formula_exactness_contract to an owner.
8. Assign every external protocol or open design choice to an owner and fail-fast policy.
9. Define public APIs at a routing level: names, inputs, outputs, contract_refs, producer_or_consumer role, and dependency notes.
10. Define api_dependency_edges and generation_order so coding can proceed without circular imports.
11. Define mode gating: which files/APIs are active for each reproduction mode and which unsupported modes must fail fast.
12. Define cross-file invariants that analyzing/coding must preserve.
13. Define unresolved issues when Stage 2 is too vague to support safe routing.

PUBLIC API RULES:
Each public API must include:

* name,
* kind,
* inputs,
* outputs,
* contract_refs,
* producer_or_consumer role,
* dependency_notes,
* contract.

Do not specify code internals.
Do not write algorithms in full.
Do not create APIs that merely expose a setter for a core artifact unless another routed API computes that artifact and a runtime consumer is specified.
Do not claim support for a mode/strategy through config values alone.
\end{tracepromptbox}

\begin{tracepromptbox}{Prompt excerpt from Per-File Contracting: anti-degradation and cross-file obligations}
SOURCE AUTHORITY:

1. Stage-3 routing information inside the compact context is authoritative for file ownership, public APIs, dependencies, mode gating, and cross-file handoffs.
2. Relevant Stage-2 alignment contracts inside the compact context are authoritative for paper semantics, core contracts, runtime boundaries, formulas, objects, targets, evaluation protocols, and forbidden downgrades.
3. Stage-1 planning fallback may be used only when explicitly included in the compact context and only to clarify a missing or weak implementation detail.
4. Implementation inference is allowed only when needed for coding feasibility and must be marked as implementation_choice.
5. If the compact context is incomplete, ambiguous, or contradictory, record an unresolved blocker or fail-fast condition. Do not guess silently.

PRIMARY PURPOSE:
Produce a per-file engineering contract that coding can directly follow.

The contract must:

* preserve all relevant core path contracts for this file;
* bind routed public APIs to concrete implementation responsibilities;
* translate handoffs into consumed/produced artifacts and call obligations;
* specify file-local algorithm steps for new methods, formulas, preprocessing, masking, metrics, or external protocols;
* restate all relevant forbidden downgrades as coding prohibitions;
* define precise fail-fast conditions for unsupported modes, missing artifacts, unavailable external protocols, or malformed inputs;
* expose shape, dtype, device, layout, prompt-target, checkpoint/state, and metric-parsing boundaries when relevant;
* prevent standard-template degradation, fake implementations, stubs, simulated objectives, or label-only substitutions;
* provide a compact handoff that can be inserted into the coding prompt.

STRICT GLOBAL RULES:

* Analyze only the target file.
* Do not create, rename, merge, split, or reassign files.
* Do not modify routed public API names unless the context explicitly marks them unresolved. If an API is too vague, keep its routed name and add required signature details.
* You may propose private helper functions only inside the same file, and only when they clarify implementation without changing public APIs.
* Do not introduce new public APIs unless required to satisfy an unresolved cross-file contract; if you do, record it under unresolved_blockers as a required routing adjustment.
* Do not replace core paper methods with generic ML defaults.
* Do not train a generative model on labels only when the contract requires full target text.
* Do not use simulated metrics, random outputs, fabricated accuracy, synthetic dummy data, mock objectives, or identity/default fallbacks for core method paths.
* Do not use a config enum, setter, wrapper, or placeholder as satisfying a core algorithm unless a routed producer and runtime consumer are identified.
* Do not claim support for unsupported modes. Unsupported core modes must be disabled or fail fast.
* Do not bury external protocol failures in downstream code. If an external protocol or released artifact is unavailable, the responsible file must fail fast before downstream training/evaluation proceeds.
* External protocols must be represented as explicit interfaces, artifact loaders, schema validators, or fail-fast blockers. Heuristic substitutes are forbidden unless the compact context explicitly allows a minimal-subset mode.
* If a routed file consumes an artifact, identify the producer and required semantics. If no producer exists, record an unresolved blocker.
* If a routed file produces an artifact, identify all consumers. If no consumer exists but the artifact is core, record an unresolved blocker.
* If a cross-file handoff is orchestrated rather than direct import, state that explicitly.
* If the compact context includes context_validation warnings, address each warning in either fail_fast_conditions, cross_file_obligations, dependency_and_import_plan, or unresolved_blockers.
  \end{tracepromptbox}

\section{Example \methodname{} Output on iTransformer}
\label{app}

This appendix provides a compact example of \methodname{}'s intermediate artifacts on the iTransformer paper. We do not include the full trajectory because the complete outputs contain long evidence records, reference registries, routing plans, and per-file contracts. Instead, we present a traceability slice centered on \texttt{src/transformer\_block.py}, since iTransformer's core mechanism depends on preserving the axis inversion: attention is computed over variate tokens. This example illustrates how \methodname{} preserves paper evidence, reconciles it into non-degradation requirements, assigns the contracts to concrete files and APIs, and anchors final code generation.

\subsection{Traceability Slice}

Table~\ref{tab:itransformer-traceability} summarizes how selected iTransformer requirements are propagated across \methodname{} stages. The table focuses on the main execution path and the transformer block slice.

\newcommand{\ownerapi}[2]{%
\begin{tabular}[t]{@{}l@{}}
\texttt{\scriptsize #1}\\
\texttt{\scriptsize #2}
\end{tabular}%
}

\begin{table*}[t]
\centering
\scriptsize
\setlength{\tabcolsep}{3.5pt}
\renewcommand{\arraystretch}{1.15}
\begin{tabular}{p{0.23\textwidth}p{0.14\textwidth}p{0.25\textwidth}p{0.25\textwidth}}
\toprule
\textbf{Paper evidence / requirement}
& \textbf{Contract}
& \textbf{Owner and API}
& \textbf{File-level obligation / code realization} \\
\midrule
Transpose raw multivariate time series from $(T \times N)$ to $(N \times T)$ before embedding.
& CR5: Data Transposition
& \ownerapi{src/data\_loader.py}{load\_and\_transpose}
& Validate input shape, perform exact transpose, preserve data order, and fail fast if output is not $(N \times T)$. \\
\midrule
Embed each variate as a token: $H^0 = \mathrm{MLP}(X)$, where $H^0 \in \mathbb{R}^{N \times D}$.
& CR1: Variate Token Embedding
& \ownerapi{src/embedding\_module.py}{embed\_variates}
& Produce one token per variate and preserve the $(N \times D)$ token layout for downstream transformer blocks. \\
\midrule
Compute self-attention over variate tokens using $\mathrm{softmax}(QK^\top / \sqrt{d_k})$.
& CR2: Inverted Self-Attention
& \ownerapi{src/transformer\_block.py}{compute\_self\_attention}
& Compute an attention map with shape $(N \times N)$ over variate tokens, not temporal positions. \\
\midrule
Apply the feed-forward network independently to each variate token.
& CR3: Feed-Forward Network
& \ownerapi{src/transformer\_block.py}{apply\_ffn}
& Apply row-wise FFN while preserving the $(N \times D)$ token matrix. \\
\midrule
Apply LayerNorm per token after transformer sublayers.
& CR4: Layer Normalization
& \ownerapi{src/transformer\_block.py}{apply\_layer\_norm}
& Normalize each token independently and preserve token identity and shape. \\
\midrule
Project final token matrix $H^L$ into forecast output $\hat{Y} \in \mathbb{R}^{S \times N}$.
& CR7: Projection to Forecast
& \ownerapi{src/projection\_module.py}{project\_to\_forecast}
& Produce forecast tensor with shape matching the ground-truth future series. \\
\bottomrule
\end{tabular}
\caption{Traceability slice for the iTransformer example. Each row shows how a paper-level requirement is turned into a reconciled contract, assigned to a concrete file/API, and localized into an implementation obligation.}
\label{tab:itransformer-traceability}
\end{table*}

\subsection{Selected Blueprint Construction and Reference Extraction Outputs}

The Blueprint Construction stage records the main reproduction scope and execution skeleton before repository architecture is fixed. For iTransformer, the blueprint identifies multivariate time series forecasting as the task, records the inverted tokenization pipeline, and separates the main method from optional ablations and analysis modules.

\begin{traceoutputbox}{Excerpt A1: Implementation blueprint execution skeleton}
1. **Data Loading & Preprocessing**
   - **Trigger:** At the start of the training/inference pipeline.
   - **Input Objects:** Raw time series dataset X ∈ ℝ^(T×N) (with provided splits and sampling frequencies; see Table 4).
   - **Operations:**
     - Load dataset following the split protocol from Appendix A.1.
     - Transpose the data from shape (T×N) to (N×T).
   - **Output Objects:** Preprocessed time series tensor X_transposed ∈ ℝ^(N×T).
   - **Artifact:** Data split information.
   - **Next Step:** MLP Embedding.
   - **Type:** paper_fact.
   - **Reproduction Mode:** train_from_released_or_precomputed_artifacts.
   - **Risk if Omitted:** Incorrect tensor shapes leading to model failure.

2. **Variate Token Embedding**
   - **Trigger:** Immediately after data loading.
   - **Input:** Preprocessed X_transposed ∈ ℝ^(N×T).
   - **Operation:**
     - For each variate (each row), apply an MLP (the embedding layer) to map a series of length T to a token vector of dimension D.
     - Produces initial token representation H^0 ∈ ℝ^(N×D).
   - **Output:** Embedded tokens H^0.
   - **Artifact:** Model weights for the embedding MLP.
   - **Next Step:** Transformer Block processing.
   - **Type:** paper_fact.
   - **Mode:** train_from_released_or_precomputed_artifacts.
   - **Risk:** Misimplementation of inversion (i.e. mixing time and variate dimensions).

3. **Transformer Block Stack**
   - **Trigger:** Run in a loop for L blocks.
   - **Input:** Token matrix H^(l) ∈ ℝ^(N×D) from the previous block or H^0.
   - **Operations for Each Block:**
     - **Self-Attention Sub-layer:**
       - Compute linearly projected queries, keys, and values from H^(l).
       - Compute attention scores across the variate tokens (resulting in an N×N matrix).
       - Apply softmax (scaled by √d_k) and produce weighted sum.
       - Add residual connection and apply Layer Normalization.
     - **Feed-Forward (FFN) Sub-layer:**
       - Apply an MLP (FFN) independently on each token.
       - Add residual connection and apply Layer Normalization.
   - **Output:** Updated token matrix H^(l+1) ∈ ℝ^(N×D).
   - **Artifact:** Model weights for attention and FFN layers.
   - **Next Step:** After L blocks, pass to the Projection Layer.
   - **Type:** paper_fact.
   - **Mode:** train_from_released_or_precomputed_artifacts.
   - **Risk:** Incorrect axis for attention/FFN may degrade performance.

4. **Projection to Forecast**
   - **Trigger:** After completion of L Transformer blocks.
   - **Input:** Final token matrix H^L ∈ ℝ^(N×D).
   - **Operation:**
     - Apply a projection MLP to each token mapping from ℝ^D to ℝ^S, where S is the forecast horizon.
     - Transpose the output such that the final prediction tensor is of shape ℝ^(S×N).
   - **Output:** Forecast tensor Ŷ ∈ ℝ^(S×N).
   - **Artifact:** Model weights for the projection MLP.
   - **Next Step:** Loss computation and training steps.
   - **Type:** paper_fact.
   - **Mode:** train_from_released_or_precomputed_artifacts.
   - **Risk:** Misalignment of dimensions which may produce incorrect forecasts.

5. **Loss Computation and Optimization**
   - **Trigger:** Within each training batch iteration.
   - **Input:** Predicted forecast Ŷ and ground truth future series Y ∈ ℝ^(S×N).
   - **Operation:**
     - Compute the L2 (MSE) loss between Ŷ and Y.
     - Backpropagate the loss.
     - Update all model parameters using the ADAM optimizer.
   - **Output:** Updated model parameters.
   - **Artifact:** Optimizer state, loss values.
   - **Type:** paper_fact.
   - **Mode:** train_from_released_or_precomputed_artifacts.
   - **Risk:** Incorrect loss implementation or optimizer hyperparameters may hinder convergence.

6. **Evaluation & Metrics Computation**
   - **Trigger:** After training (or during validation after epochs).
   - **Input:** Model predictions on the test set.
   - **Operation:**
     - Compute evaluation metrics: Mean Squared Error (MSE) and Mean Absolute Error (MAE).
     - Aggregate results over multiple forecast lengths as reported.
   - **Output:** Final reported performance metrics.
   - **Type:** paper_fact.
   - **Mode:** inference_or_evaluation_only.
   - **Risk:** Improper metric calculation can misreport model efficacy.

7. **Optional Efficient Training Strategy**
   - **Trigger:** When training on extremely high-dimensional series.
   - **Operation:**
     - Randomly sample a subset of variates per batch.
     - Use these partial variates for training while maintaining the ability to predict all variates at inference.
   - **Type:** implementation_choice.
   - **Mode:** ablation_or_analysis_only.
   - **Risk:** Without proper selection, performance differences may be misinterpreted.
\end{traceoutputbox}

The auxiliary Reference Extraction stage preserves long or exact-format benchmark materials requested by the blueprint. In this example, the long dataset description table is stored as a structured registry entry instead of being compressed into the main blueprint.

\begin{table}[ht]
\centering
\scriptsize
\setlength{\tabcolsep}{4pt}
\renewcommand{\arraystretch}{1.1}
\begin{tabular}{lcccc}
\toprule
\textbf{Dataset} & \textbf{Dim} & \textbf{Prediction length} & \textbf{Dataset size} & \textbf{Frequency} \\
\midrule
ETTh1, ETTh2 & 7 & {96, 192, 336, 720} & (8545, 2881, 2881) & Hourly \\
Weather & 21 & {96, 192, 336, 720} & (36792, 5271, 10540) & 10min \\
Traffic & 862 & {96, 192, 336, 720} & (12185, 1757, 3509) & Hourly \\
\bottomrule
\end{tabular}
\caption{Example rows from the extracted iTransformer dataset reference registry. The full table is stored in the reference registry.}
\label{tab}
\end{table}

\subsection{Selected Reconciled Contracts}

The Reconciliation stage converts evidence records into implementation contracts. The following excerpt should include only the contracts relevant to \texttt{src/transformer\_block.py}, especially CR2--CR4. This demonstrates how \methodname{} prevents the core iTransformer mechanism from being weakened into a generic transformer implementation.

\begin{traceoutputbox}{Excerpt A2: Reconciled contracts for the transformer block}
    {
      "contract_id": "CR2",
      "contract_type": "formula_exactness",
      "paper_role": "Inverted Self-Attention applied on variate tokens",
      "required_for_modes": [
        "train_from_released_or_precomputed_artifacts"
      ],
      "producer": "Attention Module",
      "consumer": "Token Aggregator within Transformer Block",
      "minimum_faithful_implementation": "Compute queries, keys, and values from token matrix H using linear projections; calculate attention as softmax((QKᵀ)/√dₖ); produce an attention map of shape (N×N) that captures multivariate correlations.",
      "forbidden_downgrades": [
        "Switching to temporal attention instead of variate attention"
      ],
      "runtime_consumption_requirement": "Attention scores must be used for weighted token updates",
      "fail_fast_if_unavailable": "Abort if attention map is not computed over variate tokens",
      "evidence": [
        "E_CR2"
      ],
      "severity_if_violated": "high",
      "routing_owner_hint": "transformer_block"
    },
    {
      "contract_id": "CR3",
      "contract_type": "formula_exactness",
      "paper_role": "Feed-Forward Network (FFN) applied independently on each variate token",
      "required_for_modes": [
        "train_from_released_or_precomputed_artifacts"
      ],
      "producer": "FFN Module within Transformer Block",
      "consumer": "Token Representation Consumer",
      "minimum_faithful_implementation": "Apply an MLP on each token (row-wise) independently to extract nonlinear representations, following the same configuration across tokens.",
      "forbidden_downgrades": [
        "Applying FFN along the wrong dimension (e.g., across time steps)"
      ],
      "runtime_consumption_requirement": "Output shape must remain (N×D)",
      "fail_fast_if_unavailable": "Abort if FFN does not process each token independently",
      "evidence": [
        "E_CR3"
      ],
      "severity_if_violated": "high",
      "routing_owner_hint": "transformer_block"
    },
    {
      "contract_id": "CR4",
      "contract_type": "formula_exactness",
      "paper_role": "Layer Normalization applied per variate token",
      "required_for_modes": [
        "train_from_released_or_precomputed_artifacts"
      ],
      "producer": "LayerNorm Module",
      "consumer": "Each Sub-layer within Transformer Block",
      "minimum_faithful_implementation": "Normalize each token vector independently as (token - mean(token)) / sqrt(var(token)) as per Equation (2)",
      "forbidden_downgrades": [
        "Normalizing across time steps or mixing variate information"
      ],
      "runtime_consumption_requirement": "Normalization output must preserve token identity",
      "fail_fast_if_unavailable": "Abort if LayerNorm does not operate per token",
      "evidence": [
        "E_CR4"
      ],
      "severity_if_violated": "high",
      "routing_owner_hint": "transformer_block"
    },

\end{traceoutputbox}

\subsection{Selected Architecture Output}

The Architecture stage assigns reconciled contracts to files and public APIs. The following excerpt should show the local ownership slice for \texttt{src/transformer\_block.py}: CR2, CR3, and CR4 are assigned to this file, and its public APIs are routed as the implementation boundary for attention, FFN, and LayerNorm.

\begin{traceoutputbox}{Excerpt A3: Ownership and handoff slice for \texttt{src/transformer\_block.py}}
    {
      "contract_id": "CR2",
      "owner_file": "src/transformer_block.py",
      "producer_apis": [
        "compute_self_attention"
      ],
      "consumer_apis": [
        "aggregate_tokens"
      ],
      "runtime_call_site": "src/transformer_block.py::compute_self_attention",
      "active_modes": [
        "train_from_released_or_precomputed_artifacts"
      ],
      "fail_fast_behavior": "Abort if attention map is not computed over variate tokens",
      "notes": "Ensure inverted self-attention is computed across variate tokens."
    },
    {
      "contract_id": "CR3",
      "owner_file": "src/transformer_block.py",
      "producer_apis": [
        "apply_ffn"
      ],
      "consumer_apis": [
        "process_tokens"
      ],
      "runtime_call_site": "src/transformer_block.py::apply_ffn",
      "active_modes": [
        "train_from_released_or_precomputed_artifacts"
      ],
      "fail_fast_behavior": "Abort if FFN does not process each token independently",
      "notes": "FFN must apply independently on each token (row-wise)."
    },
    {
      "contract_id": "CR4",
      "owner_file": "src/transformer_block.py",
      "producer_apis": [
        "apply_layer_norm"
      ],
      "consumer_apis": [
        "sub_layer_processing"
      ],
      "runtime_call_site": "src/transformer_block.py::apply_layer_norm",
      "active_modes": [
        "train_from_released_or_precomputed_artifacts"
      ],
      "fail_fast_behavior": "Abort if LayerNorm is not applied per token",
      "notes": "Layer normalization must operate on each token as specified."
    },

\end{traceoutputbox}

If space permits, the following shorter handoff excerpt may also be included to show how artifact flow is routed across files.

\begin{traceoutputbox}{Excerpt A4: Critical artifact handoff around the transformer block}
    {
      "handoff_id": "HA3",
      "artifact": "Transformer Output (H^L)",
      "source_contract_refs": [
        "CR2",
        "CR3",
        "CR4"
      ],
      "producer_file": "src/transformer_block.py",
      "consumer_file": "src/projection_module.py",
      "producer_api": "compute_self_attention / apply_ffn",
      "consumer_api": "project_to_forecast",
      "runtime_call_site": "Orchestrator in src/main.py manages artifact handoff",
      "scope": "global",
      "required_invariant": "Token structure must be preserved",
      "forbidden_implementation": "Incorrect dimension handling",
      "status": "implementable",
      "active_modes": [
        "train_from_released_or_precomputed_artifacts"
      ],
      "orchestrated_by": "src/main.py"
    },

\end{traceoutputbox}

\subsection{Per-File Contract for \texttt{src/transformer\_block.py}}

The Contracting stage localizes the global contract state into a file-level implementation contract. The excerpt below should show the concrete public interface plan and implementation recipes used to guide final code generation.

\begin{traceoutputbox}{Excerpt A5: Public interface plan and implementation recipes}
  "public_interface_plan": [
    {
      "name": "compute_self_attention",
      "kind": "function",
      "visibility": "public",
      "signature_or_inputs": [
        "embedded tokens (H^0) - tensor with shape (N×D)"
      ],
      "returns_or_outputs": [
        "attention map - tensor with shape (N×N)"
      ],
      "side_effects": [],
      "state_mutation": [],
      "contract_refs": [
        "CR2"
      ],
      "acceptance_checks": [
        "Validate attention map is computed as softmax((QKᵀ)/√dₖ) and has shape (N×N)"
      ]
    },
    {
      "name": "apply_ffn",
      "kind": "function",
      "visibility": "public",
      "signature_or_inputs": [
        "token matrix - tensor with shape (N×D)"
      ],
      "returns_or_outputs": [
        "updated tokens - tensor with shape (N×D)"
      ],
      "side_effects": [],
      "state_mutation": [],
      "contract_refs": [
        "CR3"
      ],
      "acceptance_checks": [
        "Verify that FFN is applied independently on each token and that the output shape remains (N×D)"
      ]
    },
    {
      "name": "apply_layer_norm",
      "kind": "function",
      "visibility": "public",
      "signature_or_inputs": [
        "token matrix - tensor with shape (N×D)"
      ],
      "returns_or_outputs": [
        "normalized tokens - tensor with shape (N×D)"
      ],
      "side_effects": [],
      "state_mutation": [],
      "contract_refs": [
        "CR4"
      ],
      "acceptance_checks": [
        "Check that layer normalization is applied per token and that output maintains token identity and shape"
      ]
    }
  ],
  "implementation_recipes": [
    {
      "symbol": "compute_self_attention",
      "recipe_type": "model_forward",
      "source": "stage2_contract",
      "required_steps": [
        "Validate that input embedded tokens have shape (N×D)",
        "Compute linear projections to derive queries (Q) and keys (K)",
        "Compute raw attention scores as QKᵀ, then scale by 1/√dₖ",
        "Apply softmax to obtain the attention map and verify its shape is (N×N)"
      ],
      "required_inputs": [
        "Embedded Tokens (H^0)"
      ],
      "required_outputs": [
        "attention map"
      ],
      "state_or_artifact_updates": [],
      "must_call": [],
      "must_not_call": [],
      "minimum_acceptance": "Output attention map must be computed as softmax((QKᵀ)/√dₖ) with correct shape (N×N)"
    },
    {
      "symbol": "apply_ffn",
      "recipe_type": "algorithm",
      "source": "stage2_contract",
      "required_steps": [
        "Validate that the input token matrix has shape (N×D)",
        "Apply a feed-forward network (MLP) independently to each row",
        "Ensure that the output tensor retains the shape (N×D)"
      ],
      "required_inputs": [
        "token matrix"
      ],
      "required_outputs": [
        "updated tokens"
      ],
      "state_or_artifact_updates": [],
      "must_call": [],
      "must_not_call": [],
      "minimum_acceptance": "Output updated tokens must have identical shape (N×D) with independent FFN processing per token"
    },
\end{traceoutputbox}

The next excerpt should show the compact handoff passed to the coding stage. This is usually the most compact way to demonstrate how \methodname{} converts global reasoning into a direct coding instruction for a single file.

\begin{traceoutputbox}{Excerpt A6: Compact handoff to coding}
 "compact_handoff_to_coding": {
    "one_sentence_file_goal": "Implement a Transformer Block that applies inverted self-attention, a feed-forward network, and per-token LayerNorm on embedded tokens to produce Transformer Block Output (H^L) for downstream projection.",
    "must_define": [
      "compute_self_attention",
      "apply_ffn",
      "apply_layer_norm"
    ],
    "must_call": [],
    "must_consume": [
      "Embedded Tokens (H^0)"
    ],
    "must_produce": [
      "Transformer Block Output (H^L)"
    ],
    "must_fail_fast": [
      "Fail fast if input tensor shape is not (N×D) or if computed outputs violate the invariants of CR2, CR3, or CR4"
    ],
    "forbidden_shortcuts": [
      "Do not substitute with standard temporal attention or generic FFN implementations that do not preserve per-token independence"
    ],
    "highest_priority_acceptance_checks": [
      "AC_compute_self_attention",
      "AC_apply_ffn",
      "AC_apply_layer_norm"
    ]
  }
\end{traceoutputbox}

\subsection{Generated Code Excerpt}

Finally, the Engineering stage generates code under the file-level contract. The following excerpt should include only the key functions that realize the transformer block contracts.

\begin{tracecodebox}{Excerpt A7: Generated \texttt{src/transformer\_block.py} code}
class TransformerBlock(nn.Module):
    def __init__(self, dim: int, ffn_hidden_dim: int = None) -> None:
        """
        Initializes a Transformer Block.

        Args:
            dim (int): Token dimension (D) of the input embedded tokens.
            ffn_hidden_dim (int, optional): Hidden dimension for the FFN.
                If None, defaults to 4 * dim.
        """
        super(TransformerBlock, self).__init__()
        self.dim: int = dim
        # Scale factor for attention: sqrt(dim)
        self.scale: float = math.sqrt(dim)

        # Linear projections for self-attention: query, key, and value.
        self.q_proj: nn.Linear = nn.Linear(dim, dim)
        self.k_proj: nn.Linear = nn.Linear(dim, dim)
        self.v_proj: nn.Linear = nn.Linear(dim, dim)

        # Feed-forward network (FFN)
        if ffn_hidden_dim is None:
            ffn_hidden_dim = 4 * dim
        self.ffn: nn.Sequential = nn.Sequential(
            nn.Linear(dim, ffn_hidden_dim),
            nn.ReLU(),
            nn.Linear(ffn_hidden_dim, dim)
        )

        # LayerNorm modules for each sub-layer.
        self.norm1: nn.LayerNorm = nn.LayerNorm(dim)
        self.norm2: nn.LayerNorm = nn.LayerNorm(dim)

    def compute_self_attention(self, x: torch.Tensor) -> Tuple[torch.Tensor, torch.Tensor]:
        """
        Computes self-attention over the variate tokens.

        This function projects the input tokens to queries, keys, and values. 
        Then it computes the attention scores as:
            scores = (Q @ K^T) / sqrt(dim)
        The attention map is produced by applying softmax across each row.

        Args:
            x (torch.Tensor): Input tensor of shape (N x D), where
                              N is the number of variates and D is the token dimension.

        Returns:
            Tuple[torch.Tensor, torch.Tensor]:
                - attn_output: The aggregated attention output with shape (N x D)
                - attn_map: The attention map with shape (N x N)
        """
        # x shape: (N, D)
        Q = self.q_proj(x)  # (N, D)
        K = self.k_proj(x)  # (N, D)
        V = self.v_proj(x)  # (N, D)
        # Compute raw attention scores: (N x N)
        scores = torch.matmul(Q, K.transpose(-2, -1)) / self.scale
        # Compute attention map using softmax
        attn_map = F.softmax(scores, dim=-1)  # (N, N)
        # Compute weighted output
        attn_output = torch.matmul(attn_map, V)  # (N, D)
        return attn_output, attn_map

    def apply_ffn(self, x: torch.Tensor) -> torch.Tensor:
        """
        Applies the feed-forward network (FFN) on each variate token independently.

        Args:
            x (torch.Tensor): Input tensor of shape (N x D).

        Returns:
            torch.Tensor: Output tensor of shape (N x D) after FFN.
        """
        return self.ffn(x)

    def apply_layer_norm(self, x: torch.Tensor) -> torch.Tensor:
        """
        Applies layer normalization on the input tensor.

        This static-style function computes per-token normalization: for each token (row in x),
        it subtracts the mean and divides by the standard deviation (with a small epsilon for stability).

        Args:
            x (torch.Tensor): Input tensor of shape (N x D).

        Returns:
            torch.Tensor: Normalized tensor of shape (N x D).
        """
        epsilon: float = 1e-6
        mean = x.mean(dim=-1, keepdim=True)
        var = x.var(dim=-1, unbiased=False, keepdim=True)
        return (x - mean) / torch.sqrt(var + epsilon)

    def forward(self, x: torch.Tensor) -> torch.Tensor:
        """
        Forward pass through the Transformer Block.

        The block consists of two sub-layers: a self-attention sub-layer and a feed-forward network sub-layer.
        Each sub-layer is wrapped with a residual connection and layer normalization.

        Args:
            x (torch.Tensor): Input tensor with shape (N x D).

        Returns:
            torch.Tensor: Output tensor with shape (N x D) which is the final token representation (H^L).
        """
        # Self-Attention Sub-layer
        attn_output, _ = self.compute_self_attention(x)  # attn_output: (N x D)
        # Residual connection and layer normalization (using self.norm1)
        x = self.norm1(x + attn_output)

        # Feed-Forward Sub-layer
        ffn_output = self.apply_ffn(x)  # (N x D)
        # Residual connection and layer normalization (using self.norm2)
        x = self.norm2(x + ffn_output)
        return x

\end{tracecodebox}

This example illustrates the intended role of \methodname{}'s intermediate artifacts. The paper-level axis inversion is first preserved as evidence, then converted into non-degradation requirements, routed to a concrete owner file and public APIs, localized into a file-level implementation contract, and finally reflected in code that computes attention over variate tokens with an $(N \times N)$ attention map.

\end{document}

%% file: math_commands.tex
\usepackage{amsmath,amsfonts,bm}

\def\eqref#1{equation~\ref{#1}}

\def\1{\bm{1}}

\DeclareMathAlphabet{\mathsfit}{\encodingdefault}{\sfdefault}{m}{sl}
\SetMathAlphabet{\mathsfit}{bold}{\encodingdefault}{\sfdefault}{bx}{n}



%% file: macro.tex
\definecolor{Red}{rgb}{0.6,0,0}
\definecolor{Blue}{rgb}{0,0,0.8}
\definecolor{Green}{rgb}{0,0.6,0.9}
\definecolor{airforceblue}{rgb}{0.36, 0.54, 0.66}
\definecolor{ao(english)}{rgb}{0.0, 0.5, 0.0}
\definecolor{azure(colorwheel)}{rgb}{0.0, 0.5, 1.0}
\definecolor{crimson}{rgb}{0.86, 0.08, 0.24}
\definecolor{darkcerulean}{rgb}{0.03, 0.27, 0.49}
\definecolor{cobalt}{rgb}{0.0, 0.28, 0.67}
\definecolor{rosegold}{rgb}{0.72, 0.43, 0.47}
\definecolor{orange-red}{rgb}{1.0, 0.27, 0.0}
\definecolor{mountainmeadow}{rgb}{0.19, 0.73, 0.56}
\definecolor{malachite}{rgb}{0.04, 0.85, 0.32}
\definecolor{darkblue}{rgb}{0.0, 0.0, 0.55}
\definecolor{customred}{rgb}{1, 0.85, 0.85}
\definecolor{customcitecolor}{rgb}{0.7, 0.5, 1}
\definecolor{custompink}{rgb}{0.8, 0.3, 0.3}
\definecolor{customgreen}{rgb}{0.3, 0.8, 0.3}

%% file: box_format.tex
\lstdefinestyle{traceunicode}{
    literate=
        {∈}{{\ensuremath{\in}}}1
        {∉}{{\ensuremath{\notin}}}1
        {ℝ}{{\ensuremath{\mathbb{R}}}}1
        {ℕ}{{\ensuremath{\mathbb{N}}}}1
        {×}{{\ensuremath{\times}}}1
        {≤}{{\ensuremath{\leq}}}1
        {≥}{{\ensuremath{\geq}}}1
        {≠}{{\ensuremath{\neq}}}1
        {≈}{{\ensuremath{\approx}}}1
        {→}{{\ensuremath{\rightarrow}}}1
        {←}{{\ensuremath{\leftarrow}}}1
        {∞}{{\ensuremath{\infty}}}1
        {α}{{\ensuremath{\alpha}}}1
        {β}{{\ensuremath{\beta}}}1
        {γ}{{\ensuremath{\gamma}}}1
        {λ}{{\ensuremath{\lambda}}}1
        {μ}{{\ensuremath{\mu}}}1
        {σ}{{\ensuremath{\sigma}}}1
        {θ}{{\ensuremath{\theta}}}1
        {Ŷ}{{\ensuremath{\hat{Y}}}}1
        {ᵀ}{{\ensuremath{^{\mathsf{T}}}}}1
        {√}{{\ensuremath{\surd}}}1
        {ₖ}{{\ensuremath{_{k}}}}1
}

\lstdefinestyle{traceprompt}{
    style=traceunicode,
    basicstyle=\ttfamily\tiny,
    columns=fullflexible,
    keepspaces=true,
    breaklines=true,
    breakatwhitespace=false,
    showstringspaces=false,
    upquote=true,
    tabsize=2
}

\newtcblisting{tracepromptbox}[2][]{
enhanced,
breakable,
colback=gray!3,
colframe=gray!45,
coltitle=black,
fonttitle=\bfseries\footnotesize,
title={#2},
listing only,
listing options={style=traceprompt},
left=1mm,
right=1mm,
top=0.8mm,
bottom=0.8mm,
boxsep=0.6mm,
arc=1mm,
outer arc=1mm,
#1
}

\lstdefinestyle{traceappendix}{
    style=traceunicode,
    basicstyle=\ttfamily\scriptsize,
    columns=fullflexible,
    keepspaces=true,
    breaklines=true,
    breakatwhitespace=false,
    showstringspaces=false,
    upquote=true,
    tabsize=2
}

\lstdefinestyle{traceappendixsmall}{
    style=traceunicode,
    basicstyle=\ttfamily\tiny,
    columns=fullflexible,
    keepspaces=true,
    breaklines=true,
    breakatwhitespace=false,
    showstringspaces=false,
    upquote=true,
    tabsize=2
}

\newtcblisting{traceoutputbox}[2][]{
enhanced,
breakable,
colback=gray!3,
colframe=gray!45,
coltitle=black,
fonttitle=\bfseries\footnotesize,
title={#2},
listing only,
listing options={style=traceappendixsmall},
left=1mm,
right=1mm,
top=0.8mm,
bottom=0.8mm,
boxsep=0.6mm,
arc=1mm,
outer arc=1mm,
#1
}

\newtcblisting{tracecodebox}[2][]{
enhanced,
breakable,
colback=gray!3,
colframe=gray!45,
coltitle=black,
fonttitle=\bfseries\footnotesize,
title={#2},
listing only,
listing options={style=traceappendix},
left=1mm,
right=1mm,
top=0.8mm,
bottom=0.8mm,
boxsep=0.6mm,
arc=1mm,
outer arc=1mm,
#1
}